\documentclass[lettersize,journal]{IEEEtran}

\usepackage{amsmath,amsfonts}
\usepackage{algorithmic}
\usepackage{algorithm}
\usepackage{array}
\usepackage[caption=false,font=normalsize,labelfont=sf,textfont=sf]{subfig}
\usepackage{textcomp}
\usepackage{stfloats}
\usepackage{url}
\usepackage{verbatim}
\usepackage{graphicx}
\usepackage{cite}
\usepackage{bbding}
\usepackage {amsmath}  
\usepackage{CJKutf8}
\usepackage{soul} 
\usepackage{colortbl}  
\usepackage[pagebackref,breaklinks,colorlinks,citecolor=green]{hyperref}

\usepackage{booktabs}
\usepackage{multirow}
\usepackage{bm}

\newcommand{\rev}[1]{\textcolor{black}{#1}}
\newcommand{\add}[1]{\textcolor{black}{#1}}
\newenvironment{bluerevision}{\begingroup\color{black}}{\endgroup}

\newcommand{\eg}{\textit{e.g.,~}}

\begin{document}
\title{ControlRadio: Prompt-Driven Controllable Diffusion for Cross-Modal Radio Map Generation}

\author{Kangjun Liu, Xiying Pan, Shuhang Zhang, Xiang Xiang, Ke Chen,~\IEEEmembership{Member,~IEEE},~and~Yaowei Wang 
\thanks{This work is supported in part by the Major Key Project of Pengcheng Laboratory under Grant No. PCL2025A14 and PCL2025A02, and the National Natural Science Foundation of China under Grant No. 62536003.
\textit{(K. Chen and Y. Wang are the corresponding authors of this work.)}
}
\thanks{K. Liu and K. Chen are with the Pengcheng Laboratory, China (emails: liukj@pcl.ac.cn; chenk02@pcl.ac.cn); X. Pan is with the South China University of Technology and the Pengcheng Laboratory, China (email: ftpanxiying@mail.scut.edu.cn); S. Zhang is with the Peking University, China (email: shuhangzhang@pku.edu.cn); X. Xiang is with the Huazhong University of Science and Technology, and the Pengcheng Laboratory, China (email: xex@hust.edu.cn); Y. Wang is with the Harbin Institute of Technology, Shenzhen, and also with the Pengcheng Laboratory, China (email: wangyaowei@hit.edu.cn).}
}

\markboth{Journal of \LaTeX\ Class Files,~Vol.~14, No.~8, August~2021}%
{Shell \MakeLowercase{\textit{et al.}}: A Sample Article Using IEEEtran.cls for IEEE Journals}


\maketitle

\begin{abstract}
Radio maps describe how wireless signals propagate across space and are essential for wireless communication, sensing, and network planning. However, constructing accurate radio maps traditionally requires either dense measurements or computationally expensive physical simulations, which limits scalability and real-time deployment. Recent advances in generative artificial intelligence offer a promising alternative, but existing approaches lack fine-grained control and physical consistency when applied to real-world wireless environments.
\rev{Here we present \textbf{ControlRadio}, a controllable generative framework that produces radio maps from natural-language descriptions and environmental layouts, including building structures and transmitter locations. Joint semantic and spatial conditioning enables interpretable, propagation-plausible generation, while a controlled latent prior and layout-aware conditioning improve stability and structural consistency.}
Extensive experiments demonstrate that ControlRadio achieves state-of-the-art accuracy and strong generalization across diverse urban scenarios, while reducing computation time by more than four orders of magnitude compared with conventional simulation-based methods. Such results suggest a new paradigm for scalable and controllable wireless environment modeling, with broad implications for next-generation communication systems and data-driven radio sensing.
Source codes will be made publicly available at \href{https://github.com/AkonLau/ControlRadio}{https://github.com/AkonLau/ControlRadio}.
\end{abstract}

\begin{IEEEkeywords}
Generative artificial intelligence (AI), Diffusion model, Radio map generation, Noise controller
\end{IEEEkeywords}


\section{Introduction}
\label{sec:introduction}
\IEEEPARstart{R}{adio} maps characterize the spatial distribution of wireless signal strength and spectrum usage across a physical environment, and constitute a fundamental representation for a wide range of wireless applications, including network planning~\cite{romero2022radio,romero2024theoretical},  spectrum management~\cite{bi2019engineering}, localization~\cite{yapar2023real,wang2020indoor}, environment-aware communication~\cite{ckm2021}, and path planning~\cite{mu2021intelligent,zhang2020radio}. In emerging scenarios such as dense urban network deployments, low-altitude aerial platforms, and intelligent transportation systems, accurate radio maps are increasingly essential for enabling reliable, adaptive, and autonomous wireless operations. 
Despite their importance, acquiring high-quality radio maps in practice remains a long-standing challenge. 
Direct measurement requires dense and repeated sensing campaigns, which are prohibitively costly and often infeasible at city scale~\cite{zhang2024generative}, while sparse measurements combined with interpolation or regression typically fail to capture the complex propagation phenomena present in heterogeneous environments~\cite{romero2024theoretical}.

Consequently, radio map construction has traditionally relied on physics-based electromagnetic (EM) modeling, most notably ray-tracing approaches~\cite{yun2015ray, yapar2022dataset, hoydis2023sionna}. While physically grounded, these methods face fundamental scalability and flexibility limitations. Accurate ray tracing demands detailed three-dimensional environmental models, precise material parameters, and exhaustive enumeration of multi-path interactions, resulting in extremely high compute cost in dense urban scenes. For example, generating a single $256 \times 256$ 2D radio map with meter-level resolution at one frequency, while accounting for multiple reflections, can require several minutes of computation on modern hardware. Such costs render large-scale deployment, real-time adaptation, and rapid scenario exploration impractical. Moreover, physics-based pipelines are inherently rigid, offering limited support for partial observations, dynamic environments, or task-driven constraints where semantic conditions or user intent must be explicitly incorporated. These limitations significantly hinder their applicability in next-generation, data-driven wireless systems~\cite{feng2025recent}.

\begin{figure}
\centering
\includegraphics[width=\linewidth]{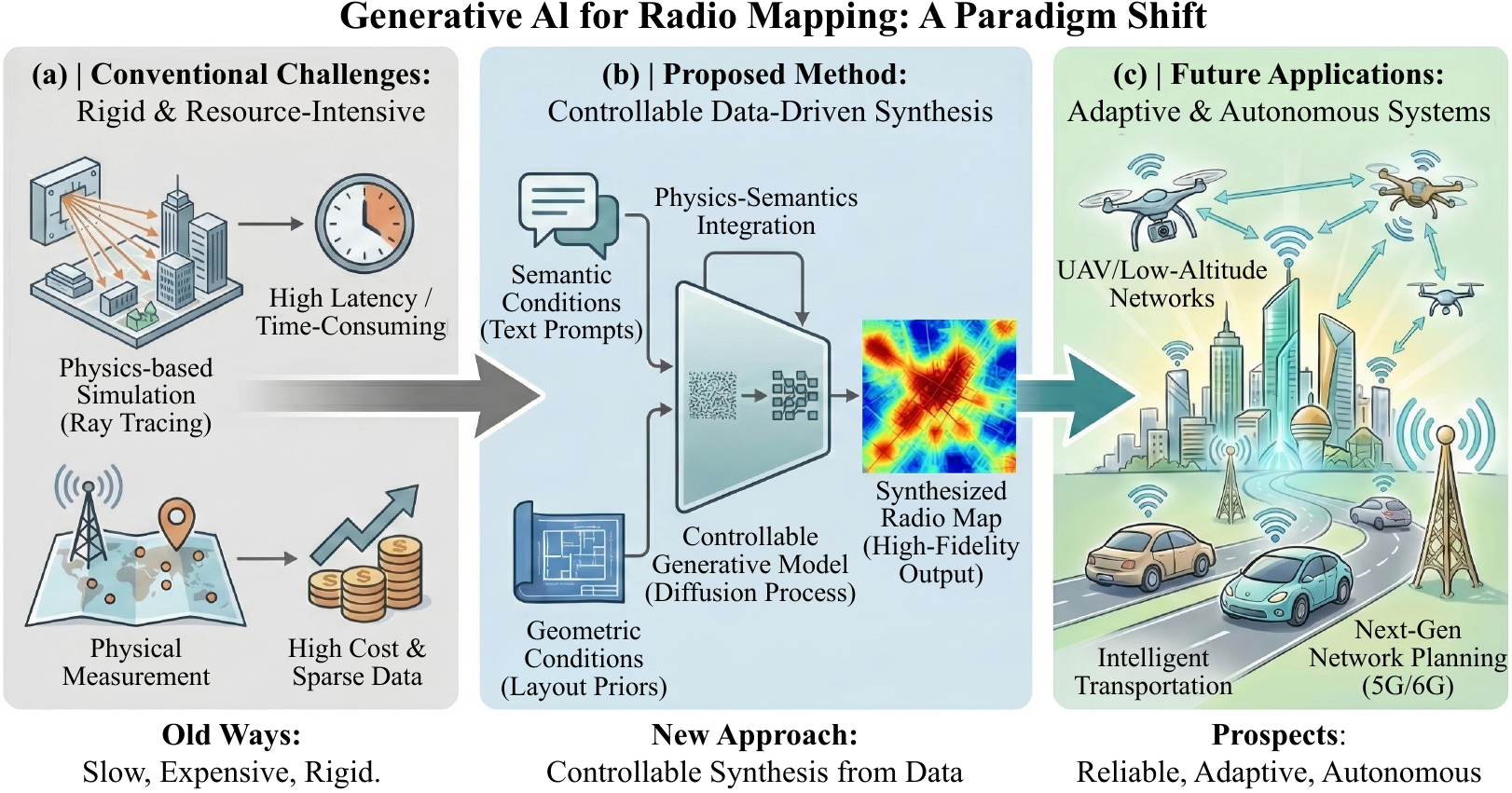}
\caption{Paradigm shift in radio map generation. The proposed controllable generative AI framework overcomes the rigidity and computational cost of conventional approaches, enabling efficient synthesis of high-fidelity radio maps to support adaptive and autonomous wireless systems.}
\label{fig:diagram}
\end{figure}

Recent advances in generative artificial intelligence, particularly diffusion probabilistic models~\cite{ho2020denoising, rombach2022high}, have demonstrated an unprecedented ability to synthesize complex, high-dimensional data under flexible multimodal control. By iteratively denoising random noise guided by text, images, or structural cues, diffusion models~\cite{yang2023diffusion} have redefined the state of the art in controllable visual content generation. These developments offer a compelling opportunity for radio map generation: rather than explicitly simulating wave propagation, generative models can learn implicit physical regularities from data and synthesize signal fields directly. However, transferring diffusion models from natural images to wireless signal modeling is far from straightforward. Radio maps are governed by strong geometry-dependent physical constraints~\cite{zhang2024physics}, where phenomena such as shadowing, diffraction, and multipath propagation are tightly coupled to environmental layout. Effective radio map generation must therefore satisfy two stringent requirements simultaneously: preserving layout-consistent spatial structure while remaining responsive to high-level semantic or task-driven conditions. Existing generative models typically address only one of these aspects, resulting in limited controllability or physically implausible outputs.

To address these challenges, we propose \textbf{ControlRadio}, a controllable diffusion-based framework that reformulates radio map generation as a cross-modal conditional synthesis problem. \rev{As illustrated in Fig.~\ref{fig:diagram}, ControlRadio unifies free-form textual prompts with explicit environmental layout constraints to generate propagation-plausible radio maps. After offline supervision generation and training, it reduces repeated online map generation from minutes to sub-second latency (typically 0.15--0.45 s per map on one GPU and 0.02--0.05 s with eight GPUs; Fig.~\ref{fig:ddim_steps}(a)). This is an amortized online-generation advantage rather than an end-to-end replacement for data acquisition or ray tracing.}
The model enables fine-grained, prompt-driven control over transmitter configurations, environmental conditions, and propagation characteristics, making it well-suited for large-scale, adaptive wireless scenarios.

As depicted in Fig.~\ref{fig:overview}, the ControlRadio framework is structured around three key technical components. First, a \textbf{Layout-Aware ControlNet} injects geometric priors, such as building morphology and transmitter placement, into the diffusion process at every denoising step, ensuring strong spatial consistency with the underlying environment. Second, we propose a novel \textbf{Noise Controller} that replaces the conventional fixed Gaussian initialization with a statistically modulated noise distribution. By explicitly controlling the mean and variance of the injected noise, this module transforms noise from a purely stochastic element into an interpretable and tunable control mechanism, substantially enhancing generation stability, controllability, and cross-sample consistency. Third, a \textbf{Decoupled Fine-tuning Strategy} leverages pretrained visual-semantic priors while efficiently adapting the model to radio-specific signal domains, enabling robust performance even under limited annotated data.
Beyond high-fidelity static map synthesis, ControlRadio supports time-varying radio map generation and facilitates downstream applications such as coverage prediction, anomaly detection, and autonomous network planning. By bridging generative AI with wireless signal modeling, ControlRadio establishes a scalable and flexible foundation for next-generation wireless environment simulation.

\vspace{0.5em}
\noindent\textbf{Our main contributions are summarized as follows:}
\begin{itemize}
    \item We introduce \textbf{ControlRadio}, the first prompt-driven and controllable diffusion framework for cross-modal radio map generation, which jointly enforces semantic intent and layout-consistent physical structure within a unified model.
    
    \item We propose a novel \textbf{Noise Controller} with statistically modulated noise injection, providing an explicit mechanism to regulate diffusion stochasticity. This design enhances generation stability, controllability, and robustness across diverse propagation conditions.
    
    \item We develop a \textbf{decoupled fine-tuning strategy} that efficiently adapts pretrained generative models to radio signal domains under limited supervision, while preserving physically meaningful spatial structures.
    
    \item \rev{Extensive simulation experiments demonstrate consistent improvements over state-of-the-art baselines in generation accuracy and controllability; propagation-aware, VAE, and geometry-only analyses further characterize physical plausibility and conditioning fidelity.} Building on the proposed framework, we further construct \textbf{TimeRadioMap}, a large-scale time-series radio map benchmark for systematic evaluation of temporal generalization beyond static reconstruction.    
\end{itemize}

The remainder of this paper is organized as follows: Section~\ref{sec:related_work} reviews related works in radio map modeling and diffusion-based generation. Section~\ref{sec:preliminaries} introduces the definition of radio map generation and the mathematical explanation of the diffusion models. Section~\ref{sec:method} presents the proposed ControlRadio framework. Section~\ref{sec:exps} provides extensive experimental validation and analysis. Finally, Section~\ref{sec:conclusion} concludes the paper and discusses future research directions.

\section{Related Work}
\label{sec:related_work}

\noindent
Our work tackles the problem of controllable radio map generation by introducing \textbf{ControlRadio}, a novel framework that integrates multi-modal conditioning and Noise Controller into a diffusion-based generative model. To contextualize our contributions, we review related literature across three core research domains: \textit{Radio Map Simulation and Generation}, \textit{Prompt-Driven Diffusion Models}, and \textit{Noise Optimization for Diffusion Generation}. These domains form the foundation upon which our proposed methodology is developed.

\vspace{0.5em}
\noindent \textbf{Radio Map Simulation and Generation.}  
Traditional radio map construction methods rely heavily on physics-based simulators or analytical propagation models, such as ray tracing~\cite{yun2015ray} or empirical path loss formulations~\cite{abhayawardhana2005comparison}. 
Although physics-based methods are interpretable and accurate when sufficient prior information is available, they often require extensive environmental modeling and high computational cost, which restricts their applicability in time-sensitive or adaptive scenarios.
To address this, recent learning-based methods~\cite{levie2021radiounet, zhang2023rme, li2024radiogat, zheng2024transformer, wang2024radiodiff} leverage deep neural networks (\eg CNNs, GNNs, transformers) to predict signal distributions from partial observations. However, most of these approaches focus on single-modality inputs (\eg occupancy maps or GPS trajectories) and lack fine-grained control over generation behavior. Some works~\cite{liu2024data, zheng2025radio} have explored semantic augmentation, but they often treat conditioning as a post-hoc bias rather than an integral part of the generative process. In contrast, ControlRadio adopts a fully conditional diffusion architecture, tightly integrating structural priors and semantic prompts throughout the generation pipeline.

\add{Recent advanced architectures broaden this landscape. RadioLAM combines propagation-based augmentation, mixture-of-experts diffusion, and physics-guided candidate selection for fine-grained 3D maps under ultra-low sampling~\cite{liu2026radiolam}; RadioDUN unfolds propagation-guided sparse recovery into an interpretable network~\cite{chen2026radiodun}; RadioAR performs coarse-to-fine autoregressive estimation~\cite{zheng2026radioar}; and CKMDiff uses a learned diffusion prior for CKM inverse problems~\cite{fu2026ckmdiff}. These methods primarily reconstruct fields from sparse or degraded numerical observations, whereas ControlRadio targets prompt- and layout-conditioned synthesis.}

\add{Beyond these learning-based reconstruction architectures, ray tracing and wireless digital twins provide physically grounded simulation and system-level scene replicas~\cite{yun2015ray,hoydis2023sionna,an2025radiotwin}. Neural radio-frequency radiance fields and 3D Gaussian-splatting methods recover continuous 3D wireless fields from sparse RF observations~\cite{zhao2023nerf2,zhang2026rf3dgs}. Their inputs, dimensionality, and objectives differ from the common 2D RadioMapSeer protocol, so including them in the same numerical comparison would confound task definition with architectural performance. ControlRadio is complementary: it amortizes controllable 2D synthesis after learning from ray-tracing references, whereas these paradigms emphasize physics-based simulation or measurement-driven 3D reconstruction.}

\vspace{0.5em}
\noindent \textbf{Prompt-Driven Diffusion Models.}  
Diffusion models have recently become a dominant paradigm in generative modeling, showing impressive performance in image synthesis~\cite{ho2020denoising, nichol2021improved}, audio generation~\cite{kong2021diffwave}, and cross-modal tasks~\cite{rombach2022high}. 
Prompt-driven diffusion incorporates external conditioning signals, most commonly natural language descriptions or spatial constraints, to steer the denoising process and produce outputs that are consistent with user intent.
ControlNet~\cite{zhang2023adding} exemplifies a key advancement by injecting structural controls (\eg edge maps or segmentation masks) into pretrained diffusion models via trainable adapters. While effective for visual domains, such designs have not been adapted to wireless signal fields, where generation must respect underlying physical constraints and spatial propagation patterns. Our framework extends this idea by incorporating building masks, transmitter placement maps, and uncertainty signals as layout-aware controls, enabling physics-informed generation under real-world signal propagation constraints.
\vspace{0.5em}
\noindent \textbf{Noise Optimization  for Diffusion Generation.}  
Noise optimization has emerged as a robust technique for enhancing generative models by refining or re-engineering random noises during either the training or inference phases. Recent works have explored a variety of noise-based strategies, including leveraging structured noise distributions like blue noise~\cite{huang2024blue} and temporally-correlated noise priors~\cite{chang2024how}, and optimizing initial noise for improved generation~\cite{guo2024initno, eyring2024reno}. Other approaches have focused on designing frameworks around specific ``golden noise"~\cite{zhou2024golden} or introducing advanced noise decomposition techniques for controllable generation, such as Factorized Diffusion~\cite{geng2024factorized} and NoiseCollage~\cite{shirakawa2024noisecollage}. Although these methods have demonstrated the potential of noise as a control signal, they often operate on a high-level, unstructured, or discrete basis. Similar to ~\cite{everaert2024exploiting}, it primarily proposes injecting signal leakage into the initial noise during inference to correct the inconsistency between training and inference in diffusion models. Our work, however, introduces a novel \textit{Noise Controller} that transforms noise injection into a tunable process by precisely controlling its statistical properties, thereby enabling a more refined control over the generative process.

\vspace{0.5em}
\noindent
In summary, our approach diverges from conventional methods by introducing a \textbf{tunable noise injection paradigm} tailored for radio signal field generation. Drawing inspiration from noise optimization techniques, ControlRadio pioneers the use of a \textbf{Noise Controller} that modulates the generative process by precisely adjusting the statistical properties of the noise. This unified framework addresses a key limitation in existing literature, enabling the generation of physically plausible radio maps that are not only \textbf{controllable via prompts} but also exhibit \textbf{superior consistency and alignment} through a novel noise-based mechanism. 

\section{Preliminaries}
\label{sec:preliminaries}

\subsection{Formulation of Radio Map Generation Problem}
A radio map provides a spatial representation of wireless signal strength or channel quality over a given geographical area, typically represented as a 2D grid of Received Signal Strength Indicator (RSSI), Signal-to-Noise Ratio (SNR), or path loss values. Formally, let $\mathcal{E}$ denote the environment, including terrain, building structures, and transmitter/receiver placements. A radio map can be expressed as a function 
\begin{equation}
R: \mathbb{R}^2 \to \mathbb{R}, \quad R(\mathbf{p}) = \text{signal metric at position } \mathbf{p},
\end{equation}
where $\mathbf{p} = (x, y)$ denotes a location coordinate in the environment.

Traditional radio map construction relies on either:  
(a) \textit{Measurement-based methods}, which interpolate sparse signal measurements collected via field surveys, \eg Kriging~\cite{boccolini2012wireless} or inverse distance weighting~\cite{lu2008adaptive}; or  
(b) \textit{Model-based methods}, which employ deterministic or stochastic propagation models, such as ray-tracing~\cite{yun2015ray} or empirical path loss models.~\cite{abhayawardhana2005comparison}  
However, these approaches face limitations in scalability, adaptability to new environments, and handling complex multipath conditions.

Recently, \textit{learning-based methods}~\cite{levie2021radiounet, zhang2023rme, wang2024radiodiff} have emerged, formulating radio map generation as a supervised or conditional generation problem. Given a set of conditions $\mathbf{c}$ (\eg building layouts, transmitter locations, transmission parameters), the objective is to learn a mapping
\begin{equation}
G_\theta(\mathbf{c}) \approx R,
\end{equation}
that can generalize to unseen layouts or configurations. Cross-modal generation further extends this formulation by incorporating high-level semantic prompts $\mathbf{t}$ (natural language descriptions), enabling flexible and human-interpretable control over the generation process.

\subsection{Diffusion Models for Conditional Generation}
Diffusion models~\cite{ho2020denoising, song2021scorebased} are a class of generative models that synthesize data by iteratively denoising a sample from a noise distribution. The forward process progressively corrupts a clean sample $\mathbf{x}_0$ with Gaussian noise over $T$ steps:
\begin{equation}
q(\mathbf{x}_t | \mathbf{x}_{t-1}) = \mathcal{N}(\sqrt{1 - \beta_t} \, \mathbf{x}_{t-1}, \beta_t \mathbf{I}),
\end{equation}
where $\beta_t$ denotes the variance schedule. The reverse process learns a parameterized denoising model $p_\theta(\mathbf{x}_{t-1} | \mathbf{x}_t)$, often implemented as a U-Net backbone, to recover $\mathbf{x}_0$.

In conditional generation, additional information $\mathbf{y}$ (\eg text embeddings, layouts) guides the denoising process via cross-attention or concatenated feature injection:
\begin{equation}
p_\theta(\mathbf{x}_{t-1} | \mathbf{x}_t, \mathbf{y}).
\end{equation}
Stable Diffusion~\cite{rombach2022high} improves scalability by operating in a latent space learned by a Variational Autoencoder (VAE), significantly reducing computational cost while preserving visual fidelity. ControlNet~\cite{zhang2023adding} further extends diffusion models with trainable condition-encoding branches that inject structured priors (\eg edges, depth maps) into intermediate layers of the U-Net, enabling fine-grained control over generation without retraining the entire backbone.

In the context of radio map generation, diffusion-based models present two notable advantages. First, they can model complex spatial patterns resulting from multipath propagation without relying on explicit physical modeling. Second, they inherently support multi-conditional inputs (\eg text prompts and environmental layouts) via modular conditioning mechanisms such as cross-attention or control networks. Despite these strengths, diffusion-based models for radio map generation still face significant challenges. 
Specifically, they struggle with controllable generation, which limits their ability to align outputs with explicit user intentions or physical constraints, and they also face training difficulties that can hinder practical deployment and performance stability.

\section{Methodology}
\label{sec:method}

To overcome the limitations of traditional radio map generation and enable controllable, semantically aligned, and structurally faithful signal synthesis, we propose \textbf{ControlRadio}, a diffusion-based generative framework jointly guided by high-level textual prompts and low-level layout constraints.
This section introduces the overall framework and details the design of prompt-guided semantic conditioning, followed by three core components: the layout-aware ControlNet, the noise controller, and the unified training and interpretability strategy. Together, these components endow ControlRadio with strong controllability, physical interpretability, and data efficiency.

\subsection{Framework Overview}

Let $\mathbf{y} \in \mathbb{R}^{H \times W}$ denote the target radio map (\eg received signal strength (RSS) or path loss) over a 2D spatial grid. Generation is conditioned on a set of multimodal inputs $\mathbf{x} = \{\mathbf{x}^{\text{text}}, \mathbf{x}^{\text{layout}}, \mathbf{x}^{\text{image}}\}$, including:

\begin{itemize}
    \item \textbf{Textual prompt} $\mathbf{x}^{\text{text}}$ describing transmission settings and environmental context (\eg frequency, power, weather);
    \item \textbf{Structural layout priors} $\mathbf{x}^{\text{layout}}$, such as building occupancy maps, transmitter positions, or semantic masks;
    \item \textbf{Optional visual context} $\mathbf{x}^{\text{image}}$, \eg aerial or satellite images for geographic reference.
\end{itemize}

\begin{figure*}[t]
\centering
\includegraphics[width=0.85\linewidth]{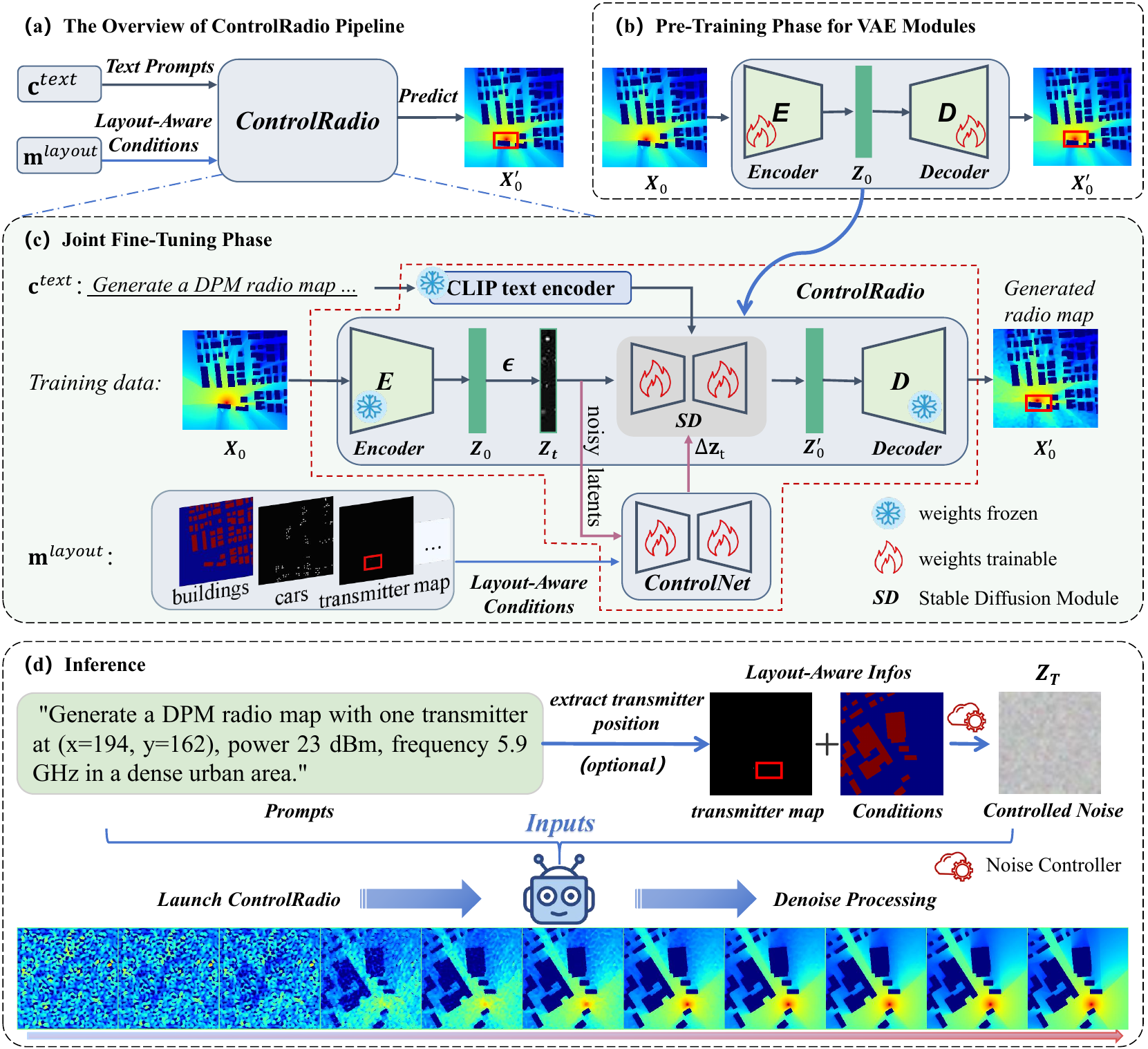}
\caption{
Overview of the proposed \textbf{ControlRadio} framework for controllable cross-modal radio map generation.
\textbf{(a)} Overall pipeline of ControlRadio. High-resolution radio maps are synthesized directly from a textual description $\mathbf{c}^{\text{text}}$ and environmental layout priors $\mathbf{m}^{\text{layout}}$, enabling flexible and interpretable signal field generation.
\textbf{(b)} Pre-training stage. A variational autoencoder (VAE) is optimized using a hybrid objective that combines $\ell_2$ reconstruction loss and perceptual loss, encouraging the latent space to preserve both fine-grained signal variations and global propagation patterns.
\textbf{(c)} Joint fine-tuning strategy. Selected modules are fine-tuned while others remain frozen, enabling effective cross-modal alignment without overfitting.
\textbf{(d)} Inference. A layout-aware Noise Controller modulates the initial noise distribution using semantic and structural cues, guiding the denoising trajectory toward physically plausible and semantically consistent radio maps.
}
\label{fig:overview}
\end{figure*}

We formulate radio map generation as a \textit{conditional denoising diffusion} problem. Starting from pure Gaussian noise $\mathbf{z}_T \sim \mathcal{N}(0, \mathbf{I})$, the model iteratively refines $\mathbf{z}_t$ to generate a physically plausible signal field $\mathbf{y}$, conditioned on multimodal inputs:

\begin{equation}
\mathbf{z}_{t-1} = \epsilon_\theta(\mathbf{z}_t, t \mid 
\mathbf{x}^{\text{text}}, \mathbf{x}^{\text{layout}}, \mathbf{x}^{\text{image}}),
\end{equation}
where $\epsilon_\theta$ is implemented as a UNet backbone augmented by a \textbf{Layout-Aware ControlNet} and a \textbf{Noise Controller}. As depicted in Fig.~\ref{fig:overview} (c), the proposed framework adopts a joint fine-tuning strategy that selectively updates task-relevant components while keeping others frozen, enabling stable adaptation to radio map generation and improving the alignment between textual semantics, environmental layouts, and propagation patterns.

\begin{figure}[t]
\centering
\includegraphics[width=\linewidth]{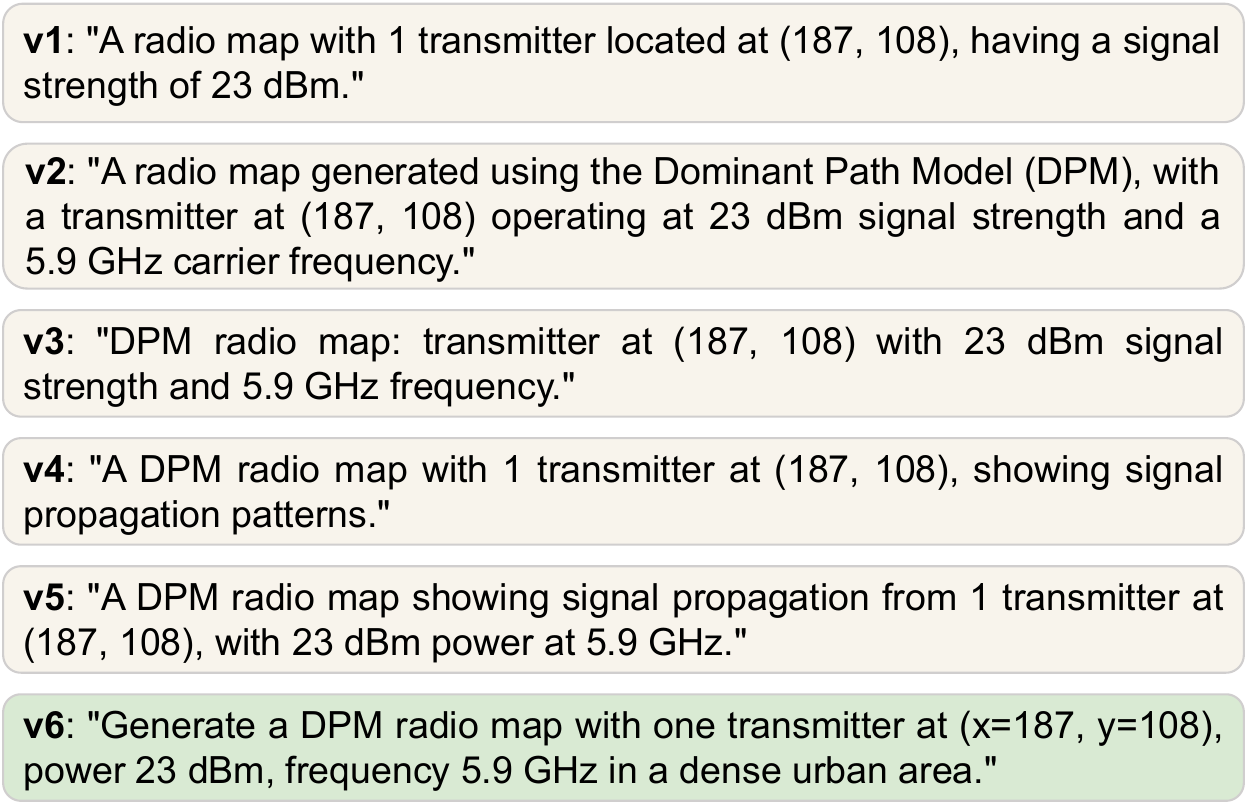}
\caption{The illustration of different prompts.}
\label{fig:prompt_types}
\vspace{-0.2cm}
\end{figure}

\subsection{Prompt-Guided Semantic Conditioning}
\label{subsec:prompt}
As illustrated in Fig.~\ref{fig:prompt_types}, we design a set of textual prompts to encode high-level semantic descriptions, which are used to condition and control the radio map generation process.
A key innovation in \textbf{ControlRadio} is the ability to generate radio maps directly from human-readable prompts. For example:

\begin{quote}
\textit{“Generate a DPM radio map with one transmitter at (x=128, y=64), power 23 dBm, frequency 5.9 GHz in a dense urban area.”}
\end{quote}
This free-form prompt is first parsed into a structured semantic vector:
\begin{equation}
\begin{split}
\mathbf{c}^{\text{text}} = \{\texttt{type}, \texttt{tx\_pos}, \texttt{power},\\ 
\texttt{frequency}, \texttt{environment}\}.
\end{split}
\end{equation}
The structured prompt is then encoded using a domain-adapted text encoder:
\begin{equation}
\mathbf{e}^{\text{text}} = E_{\text{text}}(\mathbf{c}^{\text{text}}),
\end{equation}
where $E_{\text{text}}(\cdot)$ is initialized from a pretrained CLIP~\cite{radford2021learning} encoder and partially fine-tuned to capture radio-specific semantics.  

By grounding embeddings to physically meaningful attributes (frequency, power, position, and environmental conditions), ControlRadio ensures that prompt-driven modulation respects propagation physics rather than relying solely on black-box embeddings.

\subsection{Layout-Aware ControlNet}
\label{subsec:layout}

Radio propagation is highly sensitive to environmental geometry, including obstacles, reflection surfaces, and line-of-sight availability. To incorporate structural priors, we introduce a \textbf{Layout-Aware ControlNet}, inspired by spatial conditioning in controllable image generation~\cite{zhang2023adding}.

Let $\mathbf{m}^{\text{layout}} \in \mathbb{R}^{H \times W \times C}$ represent spatial priors, including:
\begin{itemize}
    \item Building occupancy maps $\mathbf{m}^{\text{building}}$;
    \item Transmitter placement maps $\mathbf{m}^{\text{tx}}$, inferred from $\mathbf{c}^{\text{text}}$ through prompt-based semantic grounding;
    \item Optional semantic layers, \eg digital elevation models.
\end{itemize}

These are concatenated as:
\begin{equation}
\mathbf{m}^{\text{layout}} = [\mathbf{m}^{\text{building}}, \mathbf{m}^{\text{tx}}, \dots ].
\end{equation}

The ControlNet takes $\mathbf{z}_t$ and $\mathbf{m}^{\text{layout}}$ and outputs a spatial modulation term $\Delta \mathbf{z}_t$:

\begin{equation}
\Delta \mathbf{z}_t = \text{ControlNet}(\mathbf{z}_t, t, \mathbf{m}^{\text{layout}}),
\end{equation}

which is fused into the UNet via residual summation or FiLM-based feature-wise modulation~\cite{perez2018film}.  
This approach allows structural priors to be injected at all denoising steps, ensuring that the generated map naturally respects occlusions and wavefront diffusion instead of applying post-hoc masking. 

\subsection{Noise Controller for Tunable Injection}

\rev{The \textbf{Layout-Aware Noise Controller} regulates the initial latent prior rather than acting as a theoretically optimal scheduler or propagation model. Conventional inference starts from $\mathbf{z}_T\sim\mathcal{N}(0,\mathbf{I})$; we instead use}
\begin{equation}
\rev{\mathbf{z}_T\sim\mathcal{N}(\mu,\sigma^2\mathbf{I}),}
\end{equation}
\rev{and retain the standard conditional denoising trajectory thereafter. Radio maps are dominated by smooth large-scale attenuation with localized geometry-induced discontinuities. Excessive initial variance can introduce high-frequency randomness that early denoising must remove; controlling $\mu$ and $\sigma^2$ therefore supplies a simple statistical inductive bias. It does not encode an electromagnetic law.}

\rev{The parameters are selected only on the validation set and are fixed before test evaluation. The selected setting is $\mu=-0.1$ and $\sigma^2=0.001$. Fig.~\ref{fig:Parameters}(a) shows a stable neighborhood around this point rather than an isolated optimum. The controller is parameter-free, improves consistency without changing the backbone, and permits an explicit fidelity--stochasticity tradeoff. A fixed global prior may nevertheless be suboptimal under large changes in frequency, environment, or measurement domain; a condition-dependent learned prior is a natural extension.}

\subsection{Decoupled Fine-tuning Strategy}
\label{subsec:training}
To achieve efficient cross-modal alignment under limited data, we adopt a two-stage training and fine-tuning strategy:

\paragraph{Stage 1: VAE Module Pretraining.}  
Given computational and storage constraints, we first pretrain the VAE modules independently to learn stable and expressive latent representations.

\begin{itemize}
    \item \textbf{VAE Modules:} The encoder and decoder are optimized using a hybrid objective that combines $\ell_2$ reconstruction loss with perceptual loss~\cite{johnson2016perceptual}, ensuring high-fidelity reconstruction in the latent space.
    \item As illustrated in Fig.~\ref{fig:overview} (b), this pre-training stage provides a robust latent initialization that facilitates subsequent diffusion-based radio map generation.
\end{itemize}

\paragraph{Stage 2: Joint Fine-Tuning Strategy.}  
To effectively integrate structural layout conditions and semantic text information into a unified generative process, we adopt a joint fine-tuning strategy for the core components of ControlRadio, namely the \textbf{Layout-Aware ControlNet} and the \textbf{Stable Diffusion modules}, as illustrated in Fig.~\ref{fig:overview}(c). This stage aims to align spatial priors with semantic guidance while preserving the generative capacity inherited from pretrained diffusion models.

\begin{itemize}
    \item \textbf{Layout-Aware ControlNet:}
    The ControlNet branch is fully trainable during this stage and is optimized to encode environment-specific structural priors, including building morphology and transmitter layouts. By injecting layout-aware features into each denoising step, ControlNet learns to modulate spatial signal propagation patterns in a geometry-consistent manner, enabling the diffusion backbone to respect physical constraints such as occlusion, line-of-sight blocking, and spatial attenuation.

    \item \textbf{Stable Diffusion Modules:}
    The diffusion backbone is fine-tuned in a selective and controlled manner. High-level layers responsible for semantic alignment and global structure generation are gradually unfrozen, while low-level feature extractors remain partially frozen in early epochs to preserve pretrained knowledge. This strategy allows the model to adapt to radio-domain signal characteristics and prompt-conditioned generation without sacrificing training stability or overfitting under limited supervision.
\end{itemize}

Meanwhile, the \textbf{Text Encoder} and the \textbf{Noise Controller} are kept frozen during this stage to promote stable convergence and prevent overfitting.
\begin{itemize}
    \item \textbf{Text Encoder:}
    The CLIP-based text encoder is initialized from large-scale image--text pretraining corpora and remains fixed during fine-tuning, providing robust and semantically consistent prompt embeddings without introducing additional training instability.
    
    \item \textbf{Noise Controller:}
    The Noise Controller is implemented as a parameter-free module governed by a small set of predefined hyperparameters, eliminating the need for learning additional parameters while enabling explicit and interpretable control over the noise statistics.
\end{itemize}

Under this configuration, all trainable modules are jointly optimized on annotated radio map datasets using the standard diffusion objective:
\begin{equation}
\mathcal{L} =
\mathbb{E}\Big[ \big\| \epsilon_\theta(\mathbf{z}_t, t \mid \mathbf{x}) - \epsilon \big\|_2^2 \Big],
\end{equation}
which encourages the denoising network to accurately predict the injected noise while implicitly aligning prompt semantics with the generated spatial signal structure.
\rev{The two-stage strategy supports adaptation by selectively updating the layout branch or recalibrating the controlled prior. The denoising trajectory also exposes intermediate states for diagnostic inspection and interactive prompt refinement.}

In summary, ControlRadio unifies semantic prompt reasoning, layout-aware spatial control, and diffusion-based denoising within a single generative framework. \rev{This design achieves high controllability, strong agreement with simulated propagation structures, and robust generalization to unseen layouts, establishing a foundation for data-driven wireless environment modeling.}

\section{Experiment}
\label{sec:exps}
\subsection{Experimental Settings}

\subsubsection{Datasets}
To rigorously evaluate the effectiveness of the proposed \textbf{ControlRadio} framework in controllable radio map generation under realistic and diverse propagation scenarios, we utilize the \textbf{RadioMapSeer} dataset~\cite{yapar2022dataset}. \rev{This dataset is constructed using high-fidelity WinProp ray tracing~\cite{hoppe2017wave}, which models interactions such as reflection and diffraction. It provides a challenging, reproducible simulation benchmark for spectrum-aware learning in static and dynamic urban environments.}
\add{The WinProp-generated maps are used as both supervision and reference targets. Accordingly, this controlled benchmark evaluates whether ControlRadio can generate simulator-consistent radio maps for previously unseen layouts. However, these results should not be interpreted as evidence of sim-to-real generalization, material-level propagation accuracy, or equivalence to field measurements.}

RadioMapSeer consists of 701 synthetic city-scale radio maps generated from urban topologies extracted from OpenStreetMap. 
These maps cover six representative metropolitan areas, including Ankara, Berlin, Glasgow, Ljubljana, London, and Tel Aviv, encompassing a diverse range of structural densities, street layouts, and architectural morphologies.
Each map covers a $256 \times 256$ meter region with a spatial resolution of 1 meter per pixel. Building layouts are rasterized into binary morphology maps, where a value of 1 denotes structural obstruction and 0 denotes free space. Each scene contains between 50 and 150 buildings and 80 transmitter locations, randomly sampled to ensure sufficient spatial diversity and propagation variability.

To model both deterministic and uncertain wireless conditions, the dataset provides two categories of ground truth radio maps: the \textit{Static Radio Map} (SRM), representing propagation in fixed environments, and the \textit{Dynamic Radio Map} (DRM), which incorporates time-varying occlusions (\eg moving vehicles) to simulate dynamic urban contexts such as vehicular communication and mobile users. All transmitters and receivers are positioned at a uniform height of 1.5 meters, while building heights are fixed at 25 meters for consistency. The operating frequency is set to 5.9 GHz with a transmit power of 23 dBm, reflecting common vehicular communication standards.

Each data sample is converted into a set of conditioning inputs for controllable generation, including: (1) a binary morphology map encoding structural constraints, (2) a sparse transmitter mask indicating active signal sources, and (3) a natural language prompt that specifies the target generation scenario. For instance, a prompt may describe the transmitter location, power, operating frequency, and environmental context, such as “generating an IRT4 radio map with a single transmitter at $(x=98, y=121)$ operating at 23 dBm and 5.9 GHz in a dense urban area.”
\rev{In particular, the soft Uncertainty Guidance Signal (UGS) is not provided as a spatial input map but is used solely as an inference-time scalar control. UGS$=1.0$ denotes the neutral setting and introduces no additional uncertainty modulation, whereas other values are considered only in the sensitivity analysis.}

For rigorous evaluation, we partition the dataset into 500 training maps, 100 validation maps, and 101 testing maps following the ID orders, ensuring zero spatial overlap between the sets. \rev{This strict map-level split tests generalization to unseen simulated city structures and propagation conditions.}

\subsubsection{Implementation}
\begin{figure*}[htp!]
\centering
\includegraphics[width=0.95\linewidth]{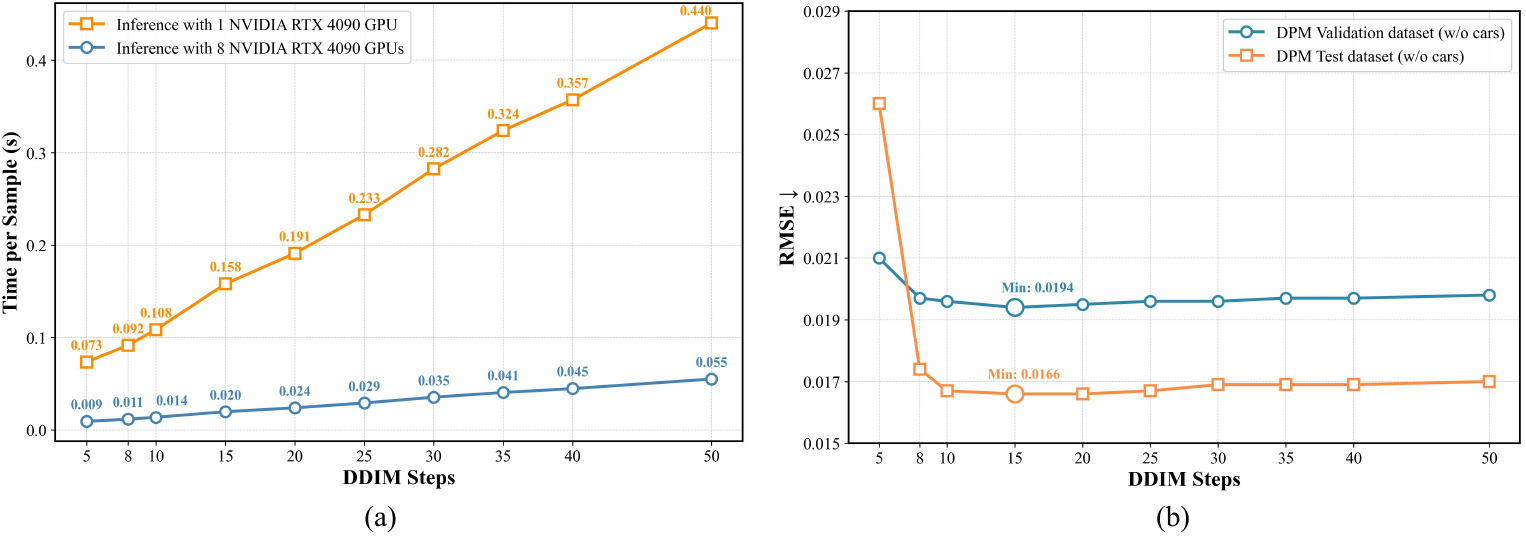}
\caption{
The influence of different inference steps.
\textbf{(a)} Inference time versus DDIM step count for ControlRadio. The computational cost scales linearly with the number of steps, and an 8-GPUs setup achieves an approximately $8\times$ speedup over a single GPU.
\textbf{(b)} RMSE versus the number of DDIM inference steps for ControlRadio. Both validation and test errors decrease rapidly and converge at around 15 steps, indicating that a small number of steps is sufficient for accurate inference.
}

\label{fig:ddim_steps}
\vspace{-0.3cm}
\end{figure*}

The \textbf{ControlRadio} framework is built upon the Stable Diffusion architecture, where we experiment with Stable Diffusion (SD) v2.1 backbone~\cite{rombach2022high}. All experiments are implemented in PyTorch and executed with mixed-precision (FP16) training on 8 NVIDIA GeForce RTX 4090 GPUs.
We initialize the model using publicly available pretrained weights and train it for 100 epochs with a learning rate of $1 \times 10^{-5}$. Optimization is carried out using the AdamW optimizer with a weight decay of 0.01. For VAE pretraining, we employ a batch size of 6 per GPU across 8 GPUs, resulting in an effective batch size of 48. During unified fine-tuning, the batch size is set to 3 per GPU across 8 GPUs. 
Unless otherwise stated, diffusion-based generation uses 50 DDIM (Denoising Diffusion Implicit Models) denoising steps. For time-series radio map generation, we adopt only 15 steps, which already achieve the optimal RMSE (Fig.~\ref{fig:ddim_steps}(b)), enabling substantially faster inference without performance degradation.

During training, we fine-tune two major components: (1) the U-Net backbone of SD, responsible for capturing joint spatial and semantic representations of radio propagation; and (2) the ControlNet module, which conditions generation on structural and transmitter-level cues. 
To balance task-specific adaptation with the preservation of pretrained knowledge, we adopt a gradual layer unfreezing strategy. Specifically, training begins by fine-tuning high-level semantic layers, while lower-level encoders are progressively unfrozen during the early training epochs, enabling stable optimization and improved generalization.
\add{All model selection is performed on the validation split. The Noise Controller and UGS settings are selected by validation sensitivity analyses and then frozen for testing. The map-level partition, preprocessing, metrics, and other common evaluation settings follow the RadioDiff-$\bm{k^2}$~\cite{wang2025radiodiff} protocol wherever applicable. The main quantitative baseline results are adopted from RadioDiff-$\bm{k^2}$ under the matched RadioMapSeer protocol, as identified in the table caption, and are not presented as re-tuned reproductions. No test map is used for training, tuning, or model selection.}

At inference time, the model takes as input a tuple comprising a binary morphological layout, a sparse transmitter mask, and a natural language prompt. \rev{No UGS map is added to this tuple; the scalar UGS is used only as an inference-time sensitivity control, with UGS$=1.0$ denoting no additional uncertainty modulation.} \rev{On one RTX 4090, generating 1,000 maps requires slightly over 7 minutes, or approximately 0.44 s per map at 50 DDIM steps. The 15-step, eight-GPU setting used for large-scale TimeRadioMap synthesis averages approximately 0.02 s per map. These numbers describe online generation after training; offline WinProp supervision and the one-time 100-epoch training run are accounted for separately in Table~\ref{tab:cost}.} \rev{For fair and consistent evaluation, all settings for ControlRadio and its ablation variants are selected on the validation split and fixed before test evaluation.}

\subsection{Evaluation Metrics}

We adopt four widely used metrics to evaluate the accuracy, fidelity, and structural integrity of the generated radio maps:

\begin{itemize}
    \item \textbf{Normalized Mean Squared Error (NMSE)}:  
    \begin{equation}
        \mathrm{NMSE} = \frac{\| \hat{R} - R \|_2^2}{\| R \|_2^2},
    \end{equation}
where $\hat{R}$ is the generated radio map, and $R$ is the ground truth. This metric reflects the normalized energy difference and penalizes magnitude errors.

    \item \textbf{Root Mean Squared Error (RMSE)}:  
    \begin{equation}
        \mathrm{RMSE} = \sqrt{\frac{1}{N} \sum_{i=1}^{N} (\hat{R}_i - R_i)^2},
    \end{equation}
where $N$ represents the number of pixels. RMSE captures the absolute pixel-wise prediction deviation.

    \item \textbf{Peak Signal-to-Noise Ratio (PSNR)}:  
    \begin{equation}
        \mathrm{PSNR} = 10 \cdot \log_{10} \left( \frac{MAX^2}{\mathrm{MSE}} \right),
    \end{equation}
where $MAX$ is the maximum possible pixel value (typically 1.0 or 255), and MSE is the mean squared error. A higher PSNR indicates better visual fidelity and lower noise.

    \item \textbf{Structural Similarity Index (SSIM)}:  
    \begin{equation}
        \mathrm{SSIM}(R, \hat{R}) = \frac{(2\mu_R \mu_{\hat{R}} + c_1)(2\sigma_{R\hat{R}} + c_2)}{(\mu_R^2 + \mu_{\hat{R}}^2 + c_1)(\sigma_R^2 + \sigma_{\hat{R}}^2 + c_2)},
    \end{equation}
where $\mu$ and $\sigma$ denote local means and variances, and $c_1, c_2$ are stabilizing constants. SSIM reflects structural and perceptual similarity.
\end{itemize}

\add{Considering that global image metrics do not establish electromagnetic correctness, we introduce three task-specific diagnostics: DAC for distance-dependent attenuation, motivated by large-scale path-loss models~\cite{3gpp38901}; LNC$_{\rm Err}$ for the signal-gap error between building-map-derived LOS/NLOS proxy regions~\cite{3gpp38901}; and BGF1 for strict pixel-overlap of high-gradient transitions, adopting the F-measure structure of boundary evaluation~\cite{martin2004boundaries}. They probe complementary macroscopic effects but not individual multipath components or Maxwell-equation compliance.}
\begin{bluerevision}
Let $R,\hat R\in[0,1]^{H\times W}$ denote the reference and generated RSS maps on the evaluation scale (stored 8-bit maps are divided by 255), and let $\mathbf p(X)$ be the radial mean profile obtained using 32 uniformly spaced map-grid distance bins. DAC combines agreement with the reference radial profile and the monotonic attenuation trend of the prediction:
\begin{equation}
\begin{aligned}
c_{\rm prof}&=\rho\!\left(\mathbf p(\hat R),\mathbf p(R)\right),\quad
c_{\rm mono}=-\rho_s\!\left(\boldsymbol{k},\mathbf p(\hat R)\right),\\
\mathrm{DAC}&=0.7c_{\rm prof}+0.3c_{\rm mono},
\end{aligned}
\label{eq:dac}
\end{equation}
where $\rho$ and $\rho_s$ denote Pearson and rank correlations, $\boldsymbol{k}$ contains the valid-bin indices, and the negative sign reflects decreasing RSS with distance. The binning and weights are fixed for all methods.

Let $\mu_{\rm L}(X)$ and $\mu_{\rm N}(X)$ be the mean values over 2-D LOS and NLOS proxy masks obtained by ray casting on the building map. The LOS/NLOS contrast error is
\begin{equation}
\begin{aligned}
\Delta_{\rm LN}(X)&=\mu_{\rm L}(X)-\mu_{\rm N}(X),\\
\mathrm{LNC}_{\rm Err}&=\left|\Delta_{\rm LN}(\hat R)-\Delta_{\rm LN}(R)\right|.
\end{aligned}
\label{eq:lnc}
\end{equation}

Finally, let $G(X)$ be the Sobel gradient magnitude, $Q_{90}$ its 90th percentile, and $i$ a pixel index. The high-gradient set and its strict zero-tolerance overlap F1 score are
\begin{equation}
\begin{aligned}
E_X&=\{i:G_i(X)\ge Q_{90}(G(X))\},\\
\mathrm{BGF1}&=\frac{2|E_{\hat R}\cap E_R|}{|E_{\hat R}|+|E_R|}.
\end{aligned}
\label{eq:bgf1}
\end{equation}
Higher DAC and BGF1 and lower LNC$_{\rm Err}$ indicate better agreement. LNC$_{\rm Err}$ is reported in normalized RSS units; tables report the per-map mean and standard deviation.
\end{bluerevision}

\subsection{Baseline Comparisons}

To evaluate the effectiveness of our proposed ControlRadio framework, we compare it against the state-of-the-art models that represent different paradigms in radio map estimation:

\begin{itemize}
    \item \textbf{RadioUNet}~\cite{levie2021radiounet}: A two-stage U-Net-based deep learning model that regresses radio propagation fields from morphological inputs and sparse transmitter annotations. It exploits spatial continuity and architectural inductive biases to generate accurate pathloss maps. While efficient and interpretable, it is less flexible under highly dynamic scenarios or when fine-grained control is required.
    
    \item \textbf{RME-GAN}~\cite{zhang2023rme}: A generative adversarial network designed for realistic radio map generation. It uses adversarial training to enforce spectral and morphological fidelity, yielding perceptually sharp results. However, the GAN structure can suffer from training instability and limited support for conditional control during generation.
    
    \item \textbf{RadioDiff}~\cite{wang2024radiodiff}: A diffusion-based generative framework tailored for high-quality radio map estimation in complex environments. It incorporates an Adaptive Fourier Transform module into the U-Net backbone to enhance spatial-frequency representation and robustness to environmental variations. The iterative denoising process enables better generalization from sparse measurements but lacks controllable semantic alignment.
    
    \item \textbf{RadioDiff-$\bm{k^2}$}~\cite{wang2025radiodiff}: A physics-informed diffusion framework for multipath-aware radio map construction, explicitly guided by the Helmholtz equation. 
    It employs a dual-diffusion design in which one model infers electromagnetic (EM) singularities, which are critical spatial features associated with negative wave numbers, while the other reconstructs the complete radio map by leveraging these singularities together with the surrounding environmental context.
    This approach bridges data-driven efficiency with physics-based EM modeling, achieving higher accuracy in complex propagation environments compared to conventional EM solvers and purely neural methods.
    
\end{itemize}

These baselines collectively cover representative generative modeling strategies, ranging from convolutional regression to diffusion and adversarial learning, enabling a comprehensive evaluation of ControlRadio in terms of controllability, accuracy, and generalization.

\subsection{Comparative Evaluation}

\begin{figure*}[t]
\centering
\includegraphics[width=0.9\linewidth]{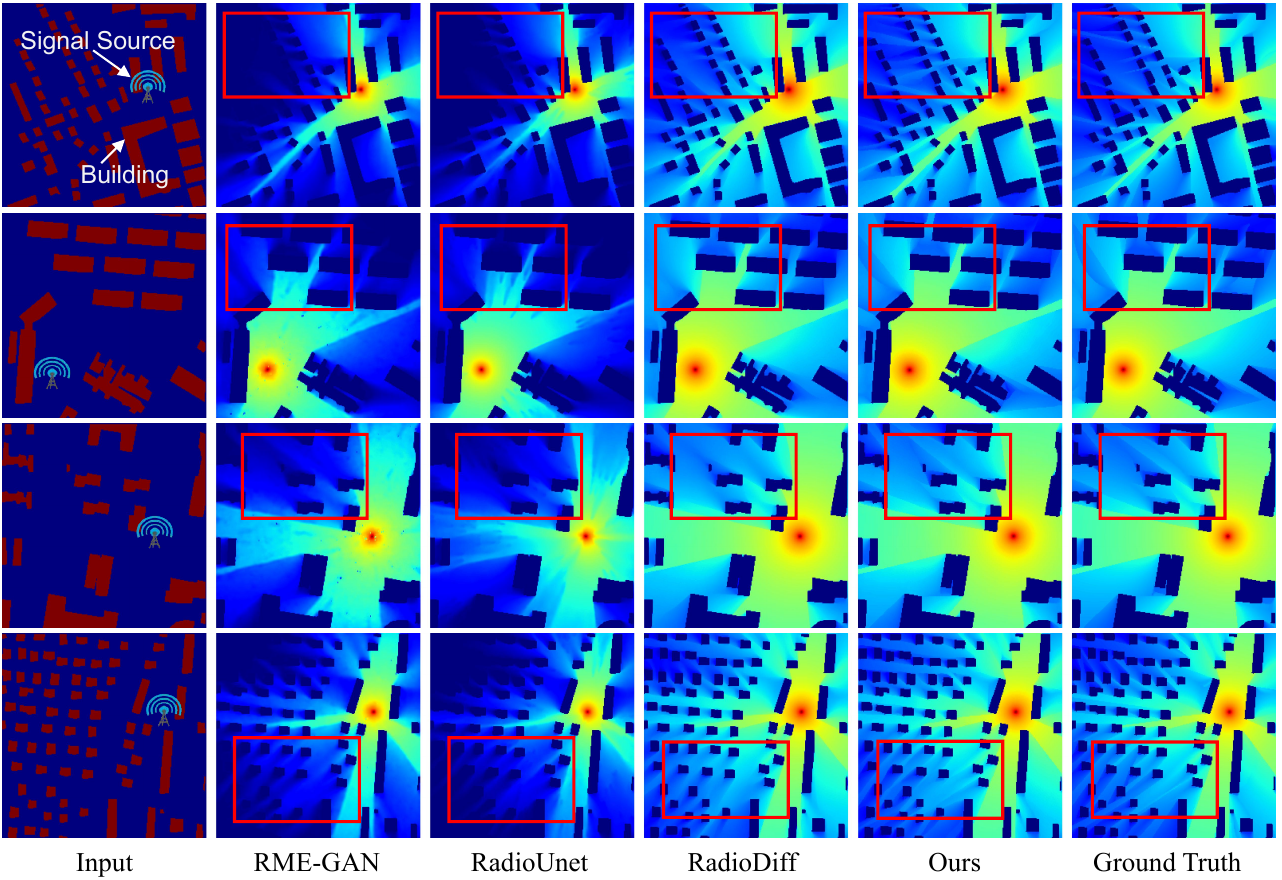}
\caption{
Qualitative comparisons on complex urban scenes. 
ControlRadio exhibits superior prompt-conditioned transformations and stronger spatial coherence under prompt v6, effectively adapting to varying environmental conditions and accurately reflecting object placement constraints. 
Qualitative results for RadioDiff-$\bm{k^2}$ are not shown, as its trained model parameters are not publicly available for reproduction and visual evaluation.
}

\label{fig:qualitative_comparisons4}
\end{figure*}

\begin{table}[t]
\centering
\caption{Quantitative comparison in terms of NMSE, RMSE, PSNR, and SSIM for different methods on the SRM (w/o cars), the DRM (DPM w/ cars), and the IRT4 ( w/o cars) simulation datasets. Baseline results are referenced from the RadioDiff-$\bm{k^2}$ paper for fair comparison.}
\label{tab:srm_dpm}
\setlength{\tabcolsep}{2mm}
\begin{tabular}{lcccc}
\toprule
\multirow{2}{*}{Methods} & \multicolumn{4}{c}{SRM (DPM w/o cars)}      \\ \cmidrule(l){2-5} 
                         & RMSE↓            & NMSE↓             & PSNR ↑            & SSIM ↑          \\ \midrule
RME-GAN                  & 0.0279           & 0.0096            & 31.35             & 0.9431 \\
RadioUNet                & 0.0266           & 0.0088            & 31.77             & 0.9466 \\
RadioDiff                  & 0.0240           & 0.0072            & 32.67             & 0.9560 \\
RadioDiff-$\bm{k^2}$             & 0.0193           & 0.0043            & 34.46             & 0.9773 \\  \midrule
ControlRadio             & \textbf{0.0166}  & \textbf{0.0024}   & \textbf{35.86}    & \textbf{0.9787} \\ \midrule\midrule

\multirow{2}{*}{Methods} & \multicolumn{4}{c}{DRM (DPM w/ cars)}                                 \\ \cmidrule(l){2-5} 
                    & RMSE↓             & NMSE↓             & PSNR ↑            & SSIM ↑          \\ \midrule
RME-GAN              & 0.0306            & 0.0115            & 30.42             & 0.9276\\
RadioUNet             & 0.0291            & 0.0107            & 30.89             & 0.9291\\
RadioDiff             & 0.0266            & 0.0090            & 31.71             & 0.9432\\
RadioDiff-$\bm{k^2}$          & 0.0208            & 0.0054            & 33.79             & 0.9704\\ \midrule
ControlRadio        & \textbf{0.0180}   & \textbf{0.0028}   & \textbf{35.29}    & \textbf{0.9759} \\ \midrule\midrule

\multirow{2}{*}{Methods} & \multicolumn{4}{c}{IRT4 ( w/o cars)}                                   \\ \cmidrule(l){2-5} 
                         & RMSE↓           & NMSE↓           & PSNR ↑           & SSIM ↑          \\ \midrule
RME-GAN                    & 0.0340          & 0.0155          & 29.74          & 0.9123          \\
RadioUNet                  & 0.0344          & 0.0159          & 29.64          & 0.9102          \\
RadioDiff                  & 0.0309          & 0.0121          & 30.44          & 0.9268          \\
RadioDiff-$\bm{k^2}$               & 0.0236          & 0.0066          & 32.68            & 0.9674          \\ \midrule
ControlRadio    & \textbf{0.0210} & \textbf{0.0040} & \textbf{33.46} & \textbf{0.9688} \\ 
\bottomrule

\end{tabular}
\end{table}

\noindent \textbf{Quantitative Comparison.} We evaluate all methods under two representative scenarios, namely \textit{Static Radio Mapping (SRM)} and \textit{Dynamic Radio Mapping (DRM)}, which correspond to stationary and time-varying environments, respectively. Performance is assessed using NMSE, RMSE, PSNR, and SSIM. As summarized in Table~\ref{tab:srm_dpm}, ControlRadio consistently outperforms all baselines on the SRM (DPM w/o cars) dataset, achieving the lowest RMSE of \textbf{0.0166} and NMSE of \textbf{0.0024}, together with the highest PSNR and SSIM. 
Compared with the strongest baseline, RadioDiff-$\bm{k^2}$, ControlRadio achieves a substantial reduction in NMSE, indicating a clear improvement in reconstruction accuracy and structural fidelity.
In more challenging DRM settings with moving vehicles, as shown in Table~\ref{tab:srm_dpm}, ControlRadio maintains robust performance despite increased temporal and environmental complexity. It achieves an RMSE of \textbf{0.0180} and an NMSE of \textbf{0.0028}, outperforming RadioDiff-$\bm{k^2}$ by a significant margin, while also improving PSNR and SSIM. This demonstrates the effectiveness of the proposed controllable diffusion framework in modeling dynamic propagation behaviors and preserving physical consistency under time-varying conditions.

Furthermore, Table~\ref{tab:srm_dpm} reports results on the IRT4 (w/o cars) dataset, which features unseen layouts and limited supervision. Even under this few-shot setting, ControlRadio achieves competitive and consistently superior performance, with the lowest RMSE and NMSE and the highest PSNR among all compared methods. These results indicate that ControlRadio generalizes well across different propagation environments and layout configurations, benefiting from its prompt-driven control mechanism and layout-aware conditioning.

\noindent \textbf{Qualitative Comparison.} We conduct comprehensive visual comparisons between ControlRadio and representative baselines, including data-driven CNN-based models, and recent diffusion-based generators. As shown in Fig.~\ref{fig:qualitative_comparisons4}, ControlRadio demonstrates superior controllability and prompt-grounded generation fidelity. Specifically, our model accurately modulates spatial field variations in response to complex multimodal prompts, while maintaining geometric consistency with layout priors. Unlike baseline models that often produce blurry or oversmoothed signal maps with poor semantic alignment, ControlRadio generates sharper, semantically coherent fields that reflect both high-level scene descriptions and low-level structural cues. Furthermore, ablation results (see Sec.~\ref{sec:ablation}) confirm the importance of both ControlNet and Noise in achieving fine-grained controllability and layout-compliant synthesis.

\subsection{Physical Fidelity Analysis}
\label{sec:physical_fidelity}
\begin{figure*}[t]
\centering
\includegraphics[width=0.98\linewidth]{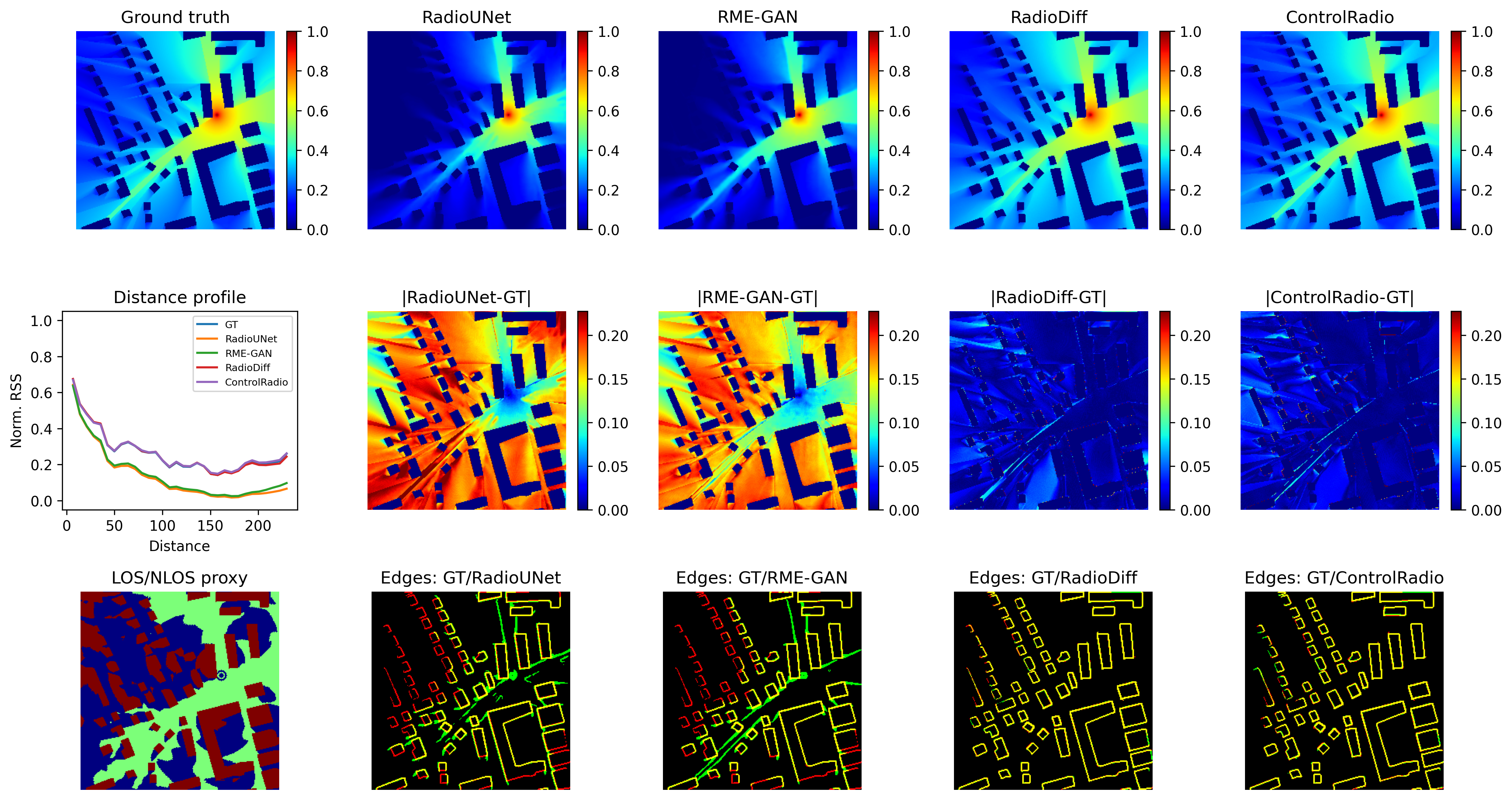}
\add{\caption{Propagation-aware diagnosis for a representative urban scene. Rows show generated maps, radial profiles and absolute errors, and the LOS/NLOS proxy with boundary overlays. Red, green, and yellow denote reference-only, prediction-only, and matched boundaries, respectively.}\label{fig:physical_diagnostics}}
\end{figure*}

\add{Table~\ref{tab:physical_diagnostics} supplements the global metrics with propagation-aware diagnostics. On all 8,080 test maps, ControlRadio obtains the lowest LNC$_{\rm Err}$ ($0.0032\pm0.0034$) and highest BGF1 ($0.8818\pm0.0575$); its DAC is comparable to RadioDiff. Although RME-GAN and RadioUNet attain higher DAC, their substantially worse contrast and boundary scores show that DAC alone does not sufficiently characterize local attenuation transitions. Fig.~\ref{fig:physical_diagnostics} likewise shows low residual error and the strongest attenuation-boundary overlap. These results support macroscopic propagation plausibility and structural consistency, not strict electromagnetic equivalence.}

\begin{table}[t]
\centering
\begingroup\color{black}
\caption{Propagation-aware diagnostics. Higher DAC and BGF1 are better; lower LNC$_{\rm Err}$ is better and is reported in normalized RSS units.}
\label{tab:physical_diagnostics}
\setlength{\tabcolsep}{1.4mm}
\scriptsize
\begin{tabular}{ccccc}
\toprule
$N$ & Method & DAC $\uparrow$ & LNC$_{\rm Err}$ $\downarrow$ & BGF1 $\uparrow$ \\
\midrule
200 & RME-GAN      & \add{0.9780$\pm$0.0280} & \add{0.0214$\pm$0.0139} & \add{0.7246$\pm$0.0736} \\
200 & RadioUNet    & \add{\textbf{0.9789$\pm$0.0260}} & \add{0.0233$\pm$0.0220} & \add{0.6662$\pm$0.0988} \\
200 & RadioDiff    & \add{0.9663$\pm$0.0457} & \add{0.0034$\pm$0.0057} & \add{0.8775$\pm$0.0571} \\
200 & ControlRadio & \add{0.9667$\pm$0.0458} & \add{\textbf{0.0031$\pm$0.0029}} & \add{\textbf{0.8884$\pm$0.0444}} \\
\midrule
8080 & RME-GAN      & 0.9767$\pm$0.0366 	& 0.0225$\pm$0.0159 		& 0.7199$\pm$0.0779 \\
8080 & RadioUNet    & \textbf{0.9770$\pm$0.0351} & 0.0250$\pm$0.0223 	& 0.6683$\pm$0.0980 \\
8080 & RadioDiff    & 0.9643$\pm$0.0520 	& 0.0035$\pm$0.0059 		& 0.8769$\pm$0.0567 \\
8080 & ControlRadio & 0.9659$\pm$0.0505 	& \textbf{0.0032$\pm$0.0034} & \textbf{0.8818$\pm$0.0575} \\

\bottomrule
\end{tabular}
\endgroup
\end{table}

\subsection{Ablation Studies}
\label{sec:ablation}
\rev{To better understand the effectiveness of each component in the ControlRadio framework, we conduct detailed ablation studies by systematically removing or modifying key modules and evaluating their impact on generation quality and controllability. We focus on four core design aspects: training data diversity, prompt conditioning, inference-time UGS sensitivity, and training strategy.}

\noindent \textbf{Effect of Training Data Diversity.} 
We investigate the impact of training data composition on the generalization performance of radio map generation. As shown in Table~\ref{tab:data_type}, models trained on a single dataset exhibit clear performance degradation when evaluated on unseen data types, indicating limited cross-domain generalization. For example, the model trained on DPM performs well on its in-domain test set but shows a substantial increase in RMSE when transferred to ITR2 and ITR4. Similar trends are observed for models trained on ITR2 or ITR4 alone, which struggle to generalize across different propagation environments.

In contrast, jointly training on a diverse mixture of datasets (DPM, ITR2, and ITR4) consistently achieves the best performance across all test sets, yielding the lowest RMSE values of \textbf{0.0166}, \textbf{0.0381}, and \textbf{0.0210}, respectively. This improvement is enabled by the use of textual prompts as a shared conditioning mechanism, which provides a unified representation of heterogeneous propagation scenarios across datasets. Such a design facilitates joint training without dataset-specific architectural modifications, allowing the model to learn more robust and transferable representations and thereby enhancing its generalization across varying environments.

\begin{table}[htp!]
\centering
\caption{RMSE comparison for radio map generation across different training and testing dataset types, evaluating cross-domain generalization performance.}
\label{tab:data_type}
\begin{tabular}{@{}lcccc@{}}
\toprule
\multirow{2}{*}{Trained Dataset}   & \multicolumn{3}{c}{Test dataset}          \\ \cmidrule(l){2-4} 
                                   & DPM             & ITR2            & ITR4   \\ \midrule
DPM                                & 0.0204     & 0.0641        & 0.0482 \\
ITR2                               & 0.0515     & 0.0448        & 0.0709 \\
ITR4                               & 0.0562 	& 0.1051 	     & 0.0536 \\
DPM, ITR2, ITR4                     & \textbf{0.0166}& \textbf{0.0381}& \textbf{0.0210}\\ \bottomrule
\end{tabular}
\end{table}

\begin{figure*}[htp!]
\centering
\includegraphics[width=0.95\linewidth]{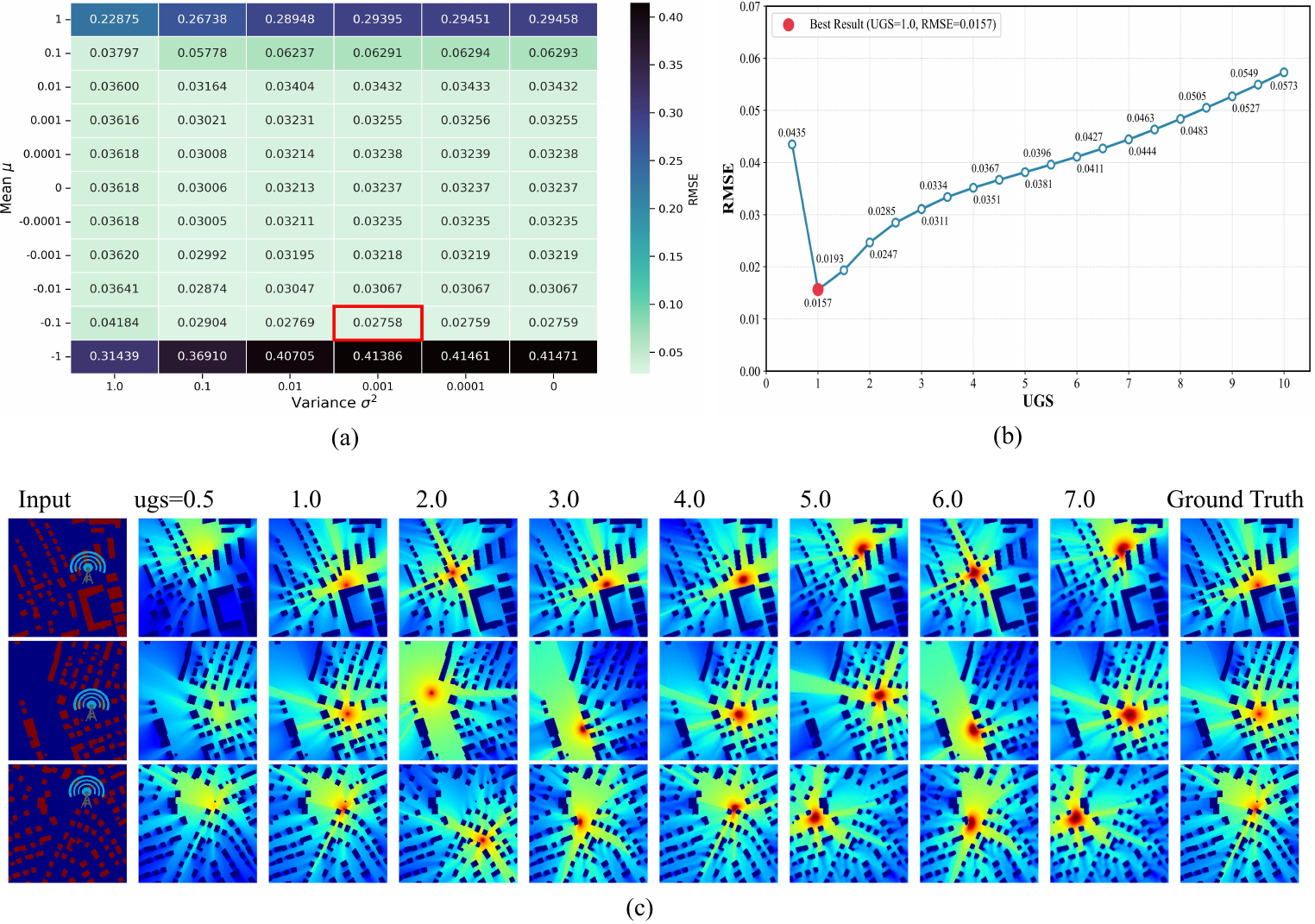}
\caption{
Hyper-parameter selection and analysis.
\textbf{(a)} RMSE comparison under different noise distribution parameters on the IRT4 (w/o cars) dataset. The best performance is obtained with noise mean $\mu=-0.1$ and variance $\sigma^2=0.001$, validating the effectiveness of controlled noise initialization.
\rev{\textbf{(b)} Sensitivity of the Uncertainty Guidance Signal (UGS) under prompt v6, evaluated on the same 80 validation samples. The neutral setting UGS$=1.0$, which adds no uncertainty term, consistently yields the lowest RMSE; deviations from this setting reduce generation fidelity.}
\rev{\textbf{(c)} Qualitative impact of UGS under prompt v6. The neutral UGS$=1.0$ setting achieves the best fidelity; UGS$=0.5$ fails to localize signal sources, while larger values introduce noticeable deviations from ground truth.}
}

\label{fig:Parameters}
\vspace{-0.3cm}
\end{figure*}

\noindent \textbf{Effect of Hyperparameter Selection.} 
We conduct a systematic hyperparameter sensitivity analysis on the ITR4 (w/o cars) simulation dataset by evaluating RMSE over a two-dimensional parameter grid. As shown in Fig.~\ref{fig:Parameters}(a), extreme parameter values on either side lead to severe performance degradation, indicating that overly strong or weak regularization destabilizes the generation process. 

\rev{The lowest validation RMSE of \textbf{0.02758} occurs at $\mu=-0.1$ and $\sigma^2=0.001$. Neighboring settings remain competitive, indicating a stable local region rather than a fragile isolated point. These parameters are chosen on validation data and then fixed for all test evaluations. This analysis supports robustness to modest perturbations within the simulated domain, while large frequency, environment, or measurement shifts may require recalibration or an adaptive prior.}

\vspace{0.5em}
\noindent \textbf{Effect of the Adaptation of the Middle Transmitter Map.} 
As reported in Table~\ref{tab:transmitter_map_noise_controller}, introducing the adaptation of the middle transmitter map leads to consistent improvements across all evaluation metrics. Specifically, the RMSE decreases from 0.0174 to \textbf{0.0166}, accompanied by a reduction in NMSE from 0.0026 to \textbf{0.0024}, indicating more accurate signal reconstruction. Meanwhile, perceptual and structural quality are further enhanced, with PSNR increasing from 35.37 to \textbf{35.86} and SSIM improving from 0.9763 to \textbf{0.9787}. These results demonstrate that adapting the middle transmitter map effectively refines intermediate feature representations, enabling better utilization of spatial signal correlations and resulting in improved reconstruction fidelity and perceptual consistency.

\begin{table}[t]
\centering
\caption{RMSE comparison illustrating the contribution of middle transmitter map adaptation and the Noise Controller. The middle transmitter map is derived from textual prompts via position extraction during data preprocessing, providing prompt-consistent spatial guidance.}
\setlength{\tabcolsep}{1mm}
\label{tab:transmitter_map_noise_controller}
\begin{tabular}{@{}cccccc@{}}
\toprule
Transmitter Map & Noise Controller & RMSE↓ & NMSE↓ & PSNR ↑ & SSIM ↑ \\ \midrule
             &          & 0.0259 	& 0.0057 	& 32.07 	& 0.9613  \\ 
$\surd$    &          &       0.0246 &       0.0052 &        32.39 &        0.9485 \\ 
    &   $\surd$       &       0.0174      &       0.0026&        35.37&        0.9763\\ 
$\surd$        &    $\surd$       & \textbf{0.0166}  & \textbf{0.0024}   & \textbf{35.86}    & \textbf{0.9787}\\ \bottomrule
\end{tabular}
\end{table}

\vspace{0.5em}
\noindent \textbf{Effect of the Noise Controller (NC).} 
As reported in Table~\ref{tab:transmitter_map_noise_controller}, introducing the Noise Controller leads to substantial performance gains across all evaluation metrics. With NC enabled, the RMSE is significantly reduced from 0.0246 to \textbf{0.0166}, and the NMSE drops from 0.0052 to \textbf{0.0024}, indicating markedly improved reconstruction accuracy. Meanwhile, perceptual and structural quality are notably enhanced, with PSNR increasing from 32.39 to \textbf{35.86} and SSIM improving from 0.9485 to \textbf{0.9787}. 

\vspace{0.5em}
\noindent \textbf{Joint Fine-tuning vs. Separate Training.}  
We compare two training strategies: (1) jointly fine-tuning the VAE Modules and the SD U-Net backbone; and (2) freezing parts of the pre-trained components and only updating partial modules. As shown in Table~\ref{tab:finetuning}, joint optimization, where both the VAE and diffusion backbone are unfrozen, yields consistently superior performance across all evaluation metrics. In particular, the RMSE is reduced to \textbf{0.0166}, compared with 0.0208 and 0.0293 when only one component is fine-tuned, and 0.0341 when all modules are frozen. Similar trends are observed for NMSE, PSNR, and SSIM, indicating not only improved reconstruction fidelity but also enhanced structural and perceptual consistency. These gains suggest that jointly adapting latent representation learning and denoising dynamics facilitates better feature alignment and task-specific specialization, which is especially critical under limited training data. Moreover, full joint fine-tuning enables more effective integration of cross-modal conditioning signals, alleviating mismatches between pre-trained priors and downstream signal characteristics, and thereby improving generalization to unseen urban layouts.

\begin{table}[t]
\centering
\caption{Effect of different fine-tuning strategies on radio map generation accuracy, measured by RMSE. Optimal performance is obtained when both the VAE and Stable Diffusion (SD) modules are fine-tuned.}
\label{tab:finetuning}
\setlength{\tabcolsep}{1mm}
\begin{tabular}{@{}lcccccc@{}}
\toprule
\multirow{6}{*}{Fine-Tuned} & \multicolumn{2}{c}{Modules} & \multirow{2}{*}{RMSE↓} & \multirow{2}{*}{NMSE↓} & \multirow{2}{*}{PSNR ↑} & \multirow{2}{*}{SSIM ↑} \\ \cmidrule(lr){2-3}
& VAE       & SD     & & & & \\ \cmidrule(l){2-7} 
&           &   & 0.0341& 0.0109& 29.68&  0.9556\\ 
&$\surd$    &   &0.0293 &0.0072 &30.81 &0.9382           \\
&           &$\surd$& 0.0208& 0.0037& 33.91&    0.9738\\
&$\surd$ &$\surd$ &\textbf{0.0166}&\textbf{0.0024}&\textbf{35.86}&\textbf{0.9787}\\
             \bottomrule
\end{tabular}
\end{table}

\add{To isolate compression error from diffusion-denoising error, we also encode and decode each reference map without diffusion sampling. Table~\ref{tab:vae_evaluation} shows RMSE below 0.0054 and SSIM above 0.9948 on all three test sets, far below the end-to-end DPM error of 0.0166. Thus, VAE compression is not the dominant bottleneck, while the joint fine-tuning results above confirm the benefit of radio-domain latent adaptation.}

\begin{table}[t]
\centering
\begingroup\color{black}
\caption{VAE-only reconstruction fidelity.}
\label{tab:vae_evaluation}
\setlength{\tabcolsep}{1.25mm}
\scriptsize
\begin{tabular}{ccccc}
\toprule
Data & RMSE $\downarrow$ & NMSE $\downarrow$ & PSNR $\uparrow$ & SSIM $\uparrow$ \\
\midrule
DPM  & \add{0.004509} & \add{0.000178} & 47.3612 & \add{0.996372} \\
IRT2 & \add{0.005317} & \add{0.000293} & 45.8893 & \add{0.995601} \\
IRT4 & \add{0.005297} & \add{0.000237} & 45.4720 & \add{0.994823} \\
\bottomrule
\end{tabular}
\endgroup
\end{table}

\begin{table}[t]
\centering
\begingroup\color{black}
\caption{Separately trained geometry-only baseline on DPM.}
\label{tab:prompt_none}
\setlength{\tabcolsep}{1.4mm}
\scriptsize
\begin{tabular}{ccccc}
\toprule
Setting & RMSE $\downarrow$ & NMSE $\downarrow$ & PSNR $\uparrow$ & SSIM $\uparrow$ \\
\midrule
Geometry only & \add{0.0337} & \add{0.0105} & 29.89 & \add{0.9340} \\
Prompt v6 & \add{\textbf{0.0166}} & \add{\textbf{0.0024}} & \textbf{35.86} & \add{\textbf{0.9787}} \\
\bottomrule
\end{tabular}
\endgroup
\end{table}

\begin{figure*}[htp!]
\centering
\includegraphics[width=0.95\linewidth]{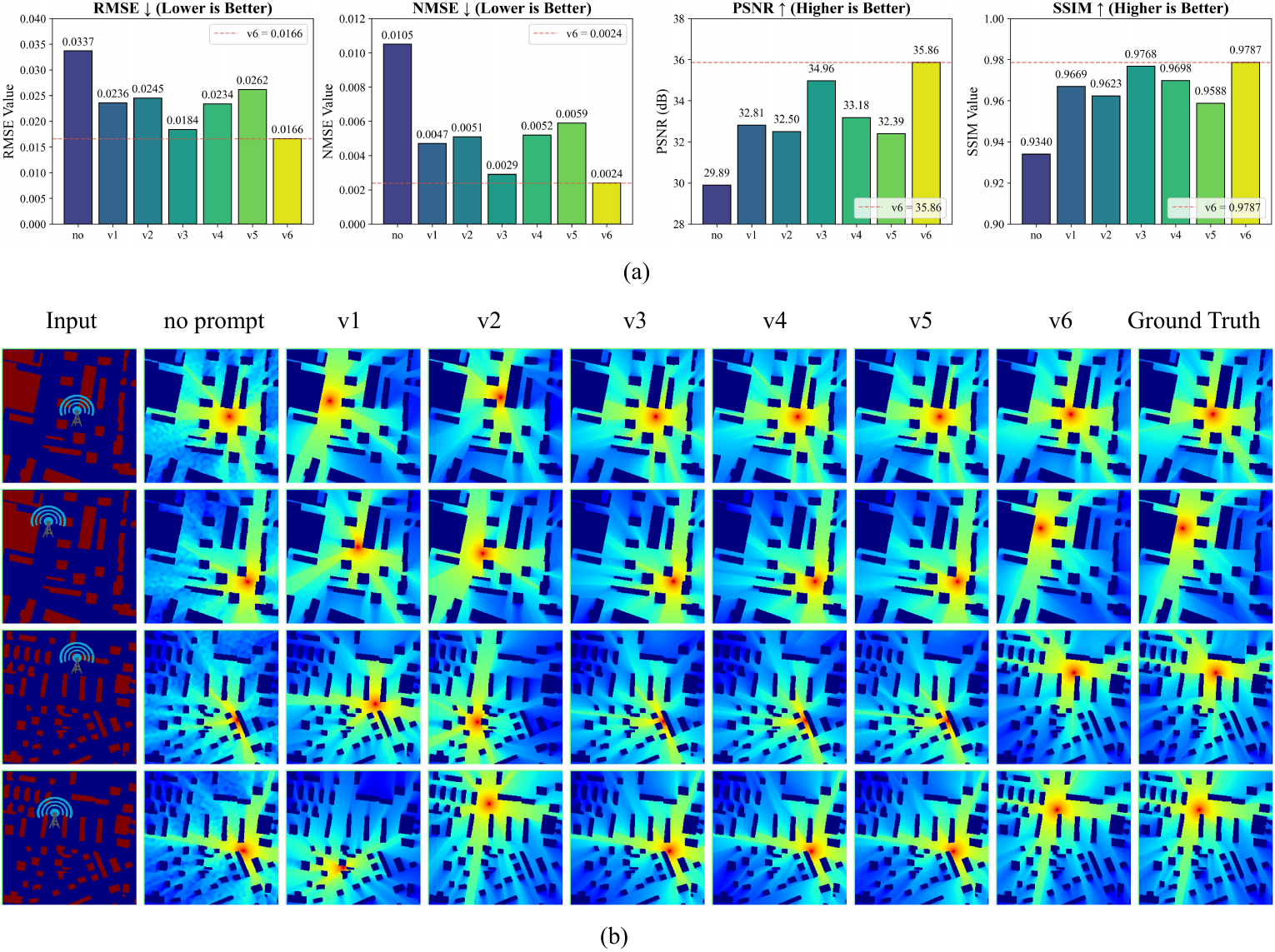}
\rev{\caption{Prompt ablation on DPM. ``No prompt'' denotes the separately trained geometry-only model, not removal of text at test time. (a) Quantitative comparison. (b) Representative maps. Prompt v6 best reproduces transmitter-centered intensity and layout-conditioned propagation patterns.}\label{fig:Ablations}}
\end{figure*}

\vspace{0.5em}
\noindent \textbf{Effect of Prompt Control.}
Figure~\ref{fig:Ablations} compares different prompt formulations to evaluate their effects on controllability and generation performance using a conventional text encoder (\eg the CLIP text encoder).
\rev{Because the model was trained without random prompt dropout, removing the text branch only at inference would cause a distribution shift. We therefore train a geometry-only baseline, denoted as ``No prompt,'' that removes the CLIP text encoder during both training and evaluation while retaining the building layout and transmitter map. Because the simulation type is specified only through text, with no equivalent nontext task token, this comparison is conducted on DPM.}

\rev{As shown in Table~\ref{tab:prompt_none} and Fig.~\ref{fig:Ablations}(a), the geometry-only baseline achieves an RMSE of 0.0337, an NMSE of 0.0105, a PSNR of 29.89~dB, and an SSIM of 0.9340. Prompt v6 improves these metrics to 0.0166, 0.0024, 35.86~dB, and 0.9787, respectively. It reduces RMSE by 50.7\% and NMSE by 77.1\%, while improving PSNR by 5.97~dB and SSIM by 0.0447. Prompt variants v1--v5 provide different levels of semantic detail, whereas v6 includes the complete condition specification: simulation type, operating frequency, transmit power, and generation instructions.}

\rev{The quantitative and qualitative results indicate that geometry provides the primary spatial prior, while text supplies complementary condition-specific information. The complete Prompt v6 achieves the best overall reconstruction fidelity and condition alignment. Thus, textual conditioning improves controllable radio map synthesis but neither replaces explicit geometric inputs nor independently guarantees electromagnetic correctness.}

\vspace{0.5em}
\noindent \textbf{Effect of Uncertainty Guidance Signal (UGS).}  
\rev{We evaluate UGS as an inference-time scalar rather than an additional spatial input. UGS$=1.0$ is the neutral setting and introduces no additional uncertainty term; values away from 1.0 are used to test sensitivity to uncertainty-related modulation. As illustrated in Fig.~\ref{fig:Parameters}(b) and Fig.~\ref{fig:Parameters}(c), UGS$=0.5$ fails to localize signal sources in several cases, whereas larger values increasingly produce spatial discontinuities or deviations from the ground truth.}

\rev{Notably, the neutral UGS$=1.0$ setting yields the lowest reconstruction error and the best structural continuity in the evaluated urban scenarios, as reflected in Fig.~\ref{fig:Parameters}(b). This result indicates that adding further uncertainty modulation does not improve this benchmark; deviations from the neutral inference configuration instead reduce fidelity.}

\begin{figure*} 
\centering
\includegraphics[width=0.9\linewidth]{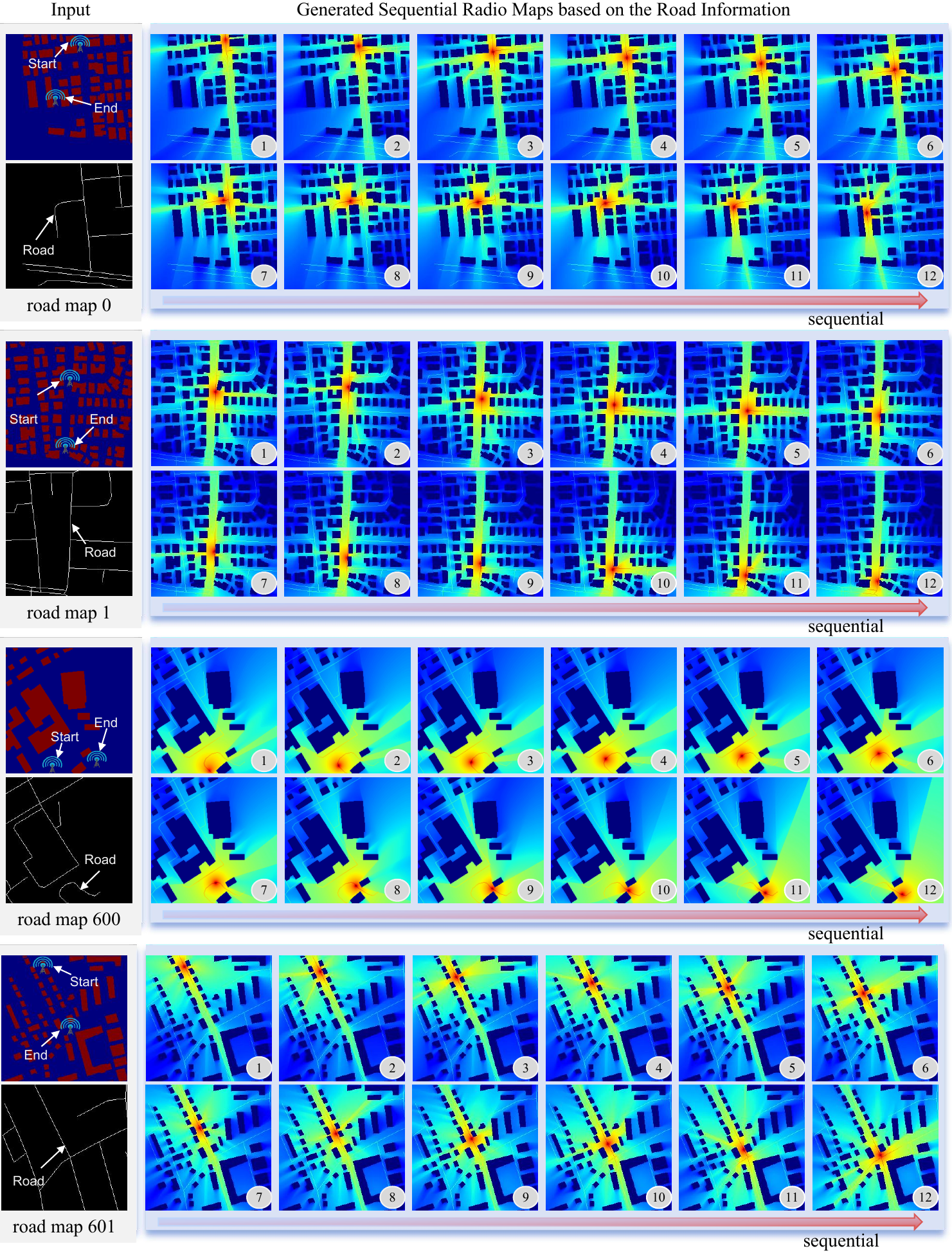}
\caption{Illustration of time-series radio map generation conditioned on road maps.
Four representative examples are shown with different building–road configurations. Road maps 0 and 1 are sampled from the training set, where the corresponding building layouts have been observed during training, while road maps 600 and 601 are drawn from the test set and correspond to previously unseen building configurations. These results demonstrate the model’s ability to generalize temporal radio map generation to novel urban layouts.}
\label{fig:road_map_generation}
\end{figure*}

\vspace{0.5em}
\noindent \textbf{Results Summary.} \rev{These ablation studies validate the necessity and complementarity of the evaluated components and provide empirical insights into the design choices behind ControlRadio. They also show that cross-modal conditioning benefits controllable synthesis, whereas the neutral UGS$=1.0$ inference configuration is more reliable than adding further uncertainty modulation on this benchmark.}

\subsection{Cost Accounting}
\add{Beyond reconstruction fidelity, a meaningful efficiency assessment requires separating offline preparation from online generation. Table~\ref{tab:cost} summarizes the corresponding lifecycle costs. ControlRadio does not eliminate the need for high-quality simulation or measurements: supervision generation and model fitting are offline costs. Its efficiency claim concerns repeated online generation after those costs have been amortized. Consequently, latency should not be interpreted as an end-to-end speedup over constructing the training corpus and training the model.}

\begin{table}[t]
\begingroup\color{black}
\centering
\caption{Offline and online cost accounting.}
\label{tab:cost}
\setlength{\tabcolsep}{1.2mm}
\scriptsize
\begin{tabular}{p{0.22\linewidth}p{0.43\linewidth}p{0.25\linewidth}}
\toprule
Stage & Reported setup & Cost role \\
\midrule
WinProp labels & Detailed ray tracing; minutes per $256^2$ map & Offline supervision \\
Training & 100 epochs, approx. 3 days, 8$\times$RTX 4090, FP16 & One-time offline \\
50-step inference & 0.44 s/map, 1$\times$RTX 4090 & Online evaluation \\
15-step inference & 0.16 s/map, 1$\times$RTX 4090 & Online evaluation \\
15-step synthesis & 0.02 s/map, 8$\times$RTX 4090 & Online large-scale \\
\bottomrule
\end{tabular}
\endgroup
\end{table}

\subsection{Time-Series Radio Map Dataset Construction}

\rev{Beyond static synthesis, ControlRadio conditions on road maps at successive time steps to generate temporally coherent radio maps without rerunning physical simulation for every frame (Fig.~\ref{fig:road_map_generation}). Road maps 0 and 1 use layouts observed during training, whereas road maps 600 and 601 use unseen test layouts. The stable outputs across both groups indicate generalization to new simulated layouts while maintaining spatial coherence and temporal continuity.}

Leveraging this property, we construct a new time-series radio map dataset, termed \textbf{TimeRadioMap}, by automatically generating radio maps over continuous time steps conditioned on road-level dynamics. 
\textbf{The TimeRadioMap dataset comprises 8,398 video sequences, each consisting of 150 frames spanning approximately 30 seconds at a spatial resolution of $256 \times 256$.}
All sequences are generated based on 701 building layouts and their corresponding road maps derived from~\cite{yapar2022dataset}.
\rev{After the offline supervision and training costs, TimeRadioMap can be synthesized with substantially lower per-map online latency than repeated ray tracing, providing a scalable benchmark for time-dependent radio map reconstruction, prediction, and low-altitude sensing applications.} 
\textbf{Notably, using ControlRadio, generating such a dataset comprising approximately 1.26 million radio maps requires only about 8 hours on a single platform equipped with 8 NVIDIA GeForce RTX 4090 GPUs.}
This dataset further enables systematic investigation of temporal generalization and controllable radio environment modeling, which remains largely unexplored in existing radio map benchmarks.

\subsection{Discussion and Limitations}
\label{sec:discussion}
\add{The present evidence is limited to simulation benchmarks. The adopted diagnostics assess distance-dependent attenuation, LOS/NLOS contrast, and blockage boundaries; however, scalar radio maps cannot verify individual multipath components, diffraction mechanisms, or material parameters. Practical deployment therefore requires measurement-based evaluation and calibration. A direct approach is to fine-tune or adapt the model using sparse drive-test samples while retaining the simulation-trained prior. Extending the present evaluation beyond 2D scalar radio maps may benefit from complementary paradigms, including ray tracing, neural radiation fields, and 3D Gaussian-splatting-based wireless reconstruction.}

\add{Sparse observations could further support online correction. Coordinate--value statements such as ``the reference level at $(x,y)$ is $-90$ dBm'' can be serialized in the prompt and injected through cross-attention. Because text alone may not enforce exact point-wise values, a reliable implementation should combine it with an auxiliary sparse-observation map and an observation-consistency loss or a sampling-time data-consistency step. This design would enable natural-language interaction while imposing hard or near-hard measurement constraints. Since it changes the conditioning interface and requires retraining, measurement-guided synthesis is considered future work rather than a current capability.}
\section{Conclusion}
\label{sec:conclusion}

\rev{In this study, we presented \textbf{ControlRadio}, a controllable diffusion framework that jointly uses semantic prompts and structural layouts. Its layout-aware ControlNet enforces spatial conditioning, its Noise Controller supplies a tunable initial latent prior, and joint adaptation aligns pretrained components with radio-map statistics. On RadioMapSeer simulation benchmarks, ControlRadio improves reconstruction and structural metrics over the compared 2D baselines. Propagation-aware diagnostics support macroscopic plausibility, VAE-only reconstruction shows limited compression loss, and the consistently trained geometry-only comparison demonstrates complementary value from text beyond geometry.}

\rev{These findings establish controllable, amortized synthesis of simulator-consistent radio maps rather than measurement-validated electromagnetic equivalence. Future work will evaluate sim-to-real transfer, learn condition-dependent noise priors, and incorporate sparse RF observations through data-consistent measurement guidance. Richer LiDAR and RF-native inputs, accelerated sampling, and continuous 3D representations may further connect ControlRadio to practical wireless digital twins.}

\vspace{0.1cm}

\bibliographystyle{IEEEtran}
\bibliography{reference}

\vfill

\end{document}